%% file: collas2024_conference.tex
\documentclass{article} 
\usepackage{collas2024_conference,times}
\usepackage{easyReview}

\usepackage{subfig}

\input{math_commands.tex}

\usepackage{booktabs}
\usepackage{afterpage}
\usepackage{hyperref}
\hypersetup{
    colorlinks=true,
    linkcolor=red,
    filecolor=magenta,
    urlcolor=blue,
    citecolor=purple,
    pdftitle={Overleaf Example},
    pdfpagemode=FullScreen,
    }
\usepackage{arydshln}
\usepackage{tabularray}
\usepackage[inkscapeformat=png]{svg}
\usepackage{enumitem}
\usepackage{multirow}
\usepackage{subfig}

\usepackage{array}
\newcolumntype{L}[1]{>{\raggedright\let\newline\\\arraybackslash\hspace{0pt}}m{#1}}
\newcolumntype{C}[1]{>{\centering\let\newline\\\arraybackslash\hspace{0pt}}m{#1}}
\newcolumntype{R}[1]{>{\raggedleft\let\newline\\\arraybackslash\hspace{0pt}}m{#1}}

\usepackage[makeroom]{cancel}
\usepackage{physics}
\usepackage[linesnumbered,ruled]{algorithm2e}
\title{A Contrastive Symmetric Forward-Forward Algorithm (SFFA) for Continual Learning Tasks}

\collasfinalcopy 

\author{Erik B. Terres-Escudero\textsuperscript{1}, Javier Del Ser\textsuperscript{2,3}, Pablo Garcia-Bringas\textsuperscript{1} \\
\textsuperscript{1} Department of Engineering, DeustoTech, University of Deusto, Spain\\
\textsuperscript{2} TECNALIA, Basque Research \& Technology Alliance (BRTA), 48160 Derio, Spain \\
\textsuperscript{3} University of the Basque Country (UPV/EHU), 48013 Bilbao, Spain\\
\texttt{e.terres@deusto.es}
}

\newcommand{\accdiff}[1]{\hspace{3pt} {\small \textcolor{red}{#1\,\textdownarrow} }}
\newcommand{\accdiffb}[1]{\hspace{3pt} {\small \textcolor{red}{\textbf{#1}\,\textdownarrow} }}

\preprintcopy 

\begin{document}

\maketitle

\begin{abstract}
The so-called Forward-Forward Algorithm (FFA) has recently gained momentum as an alternative to the conventional back-propagation algorithm for neural network learning, yielding competitive performance across various modeling tasks. By replacing the backward pass of gradient back-propagation with two contrastive forward passes, the FFA avoids several shortcomings undergone by its predecessor (e.g., vanishing/exploding gradient) by enabling layer-wise training heuristics. In classification tasks, this contrastive method has been proven to effectively create a latent sparse representation of the input data, ultimately favoring discriminability. However, FFA exhibits an inherent asymmetric gradient behavior due to an imbalanced loss function between positive and negative data, adversely impacting on the model's generalization capabilities and leading to an accuracy degradation. To address this issue, this work proposes the Symmetric Forward-Forward Algorithm (SFFA), a novel modification of the original FFA which partitions each layer into positive and negative neurons. This allows the local fitness function to be defined as the ratio between the activation of positive neurons and the overall layer activity, resulting in a symmetric loss landscape during the training phase. To evaluate the enhanced convergence of our method, we conduct several experiments using multiple image classification benchmarks, comparing the accuracy of models trained with SFFA to those trained with its FFA counterpart. As a byproduct of this reformulation, we explore the advantages of using a layer-wise training algorithm for Continual Learning (CL) tasks. The specialization of neurons and the sparsity of their activations induced by layer-wise training algorithms enable efficient CL strategies that incorporate new knowledge (classes) into the neural network, while preventing catastrophic forgetting of previously learned concepts. Experiments in three CL scenarios (Class, Domain, and Task Incremental) using multiple well-known CL techniques (EWC, SI, MAS, Replay and GEM) are discussed to analyze the differences between our SFFA model and a model trained using back-propagation. Our results demonstrate that the herein proposed SFFA achieves competitive levels of accuracy when compared to the off-the-shelf FFA, maintaining sparse latent activity, and resulting in a more precise goodness function. Our findings support the effectiveness of SFFA in CL tasks, highlighting its natural complementarity with techniques devised for this modeling paradigm.
\end{abstract}

\section{Introduction}
\label{sec:introduction}

The need for highly robust AI models capable of adapting to unexpected events (e.g. out-of-distribution data or adversarial attacks) has emerged as a crucial requirement for safe systems that rely on this technology. This has become particularly challenging in open-world learning environments \citep{parmar2023open}, in which the contextual variability and the interaction of the model with the real world can give rise to unintended consequences or harm for the end user. This non-stationary nature poses a challenge for the adaptability of AI models, compelling them to update their knowledge steadily over time. Unfortunately, in doing so AI models face the so-called catastrophic forgetting effect, by which the acquisition of new concepts becomes detrimental to previously acquired knowledge. This effect is known to heavily affect models trained using the back-propagation algorithm, which has become the default training algorithm for modern systems based on neural computation \citep{goodfellow2013empirical}. 

In response to this challenge, the last decade has witnessed the emergence and progressive maturity of Continual Learning (CL), which strives to endow AI models with the capability to assimilate novel concepts without compromising the performance gained in previously seen tasks. Within this framework, various scenarios have been proposed to find a proper balance between plasticity and stability, with Class Incremental Learning (IL), Domain IL, and Task IL outstanding as the most widely CL tasks tackled in the literature. Addressing these scenarios involves developing robust learning heuristics capable of overcoming the drawbacks associated with gradient-based learning algorithms beyond the aforementioned catastrophic forgetting effect. Other downsides such as vanishing/exploding gradients or the computational effort make models trained via gradient back-propagation unsuitable in the current CL landscape.

In this regard, a recent paper from \citet{hinton2022forward} introduced the Forward-Forward Algorithm (FFA), a novel training algorithm for neural networks. This algorithm eliminates the need for the back-propagation step in the training process, replacing it with a contrastive process involving two forward passes. By omitting this back-propagation step, FFA allows for the use of isolated layer-wise goodness functions, reducing the impact of several of the aforementioned shortcomings, including vanishing or exploding gradients. A subsequent study by \citet{tosato2023emergent} revealed that during training, this approach generates a sparse representation of the data, exposing that neurons trained via FFA become highly specialized in distinguishing specific classes. Despite demonstrating competitive accuracy with the back-propagation algorithm, the applicability of FFA to CL scenarios is still unexplored. Our hypothesis is that the representational properties featured by FFA holds potential for incorporating new knowledge to neural networks used for IL tasks, while providing at the same time the benefits inherent brought by this algorithm compared to gradient back-propagation.

In this study we address this hypothesis, introducing a novel variant of the naive FFA coined as Symmetric Forward-Forward Algorithm (SFFA). This method introduces a novel goodness function designed to overcome the asymmetry observed within the gradient of negative samples, a factor known to affect severely the generalization capability of the learned model. By partitioning each layer into positive and negative neuron sets and formulating the goodness function as the ratio of their activations, we produce a balanced loss landscape that enhances the convergence of the model during training. Furthermore, we show that by effect of SFFA, the number of updates on specialized neurons can be reduced, mitigating the catastrophic forgetting in IL tasks. We evaluate SSFA over several image classification tasks to compare its performance against FFA and gradient back-propagation. Additionally, we benchmark conventional CL techniques trained with SFFA over three IL scenarios (Class IL, Domain IL, and Task IL). Our experiments reveal that CL techniques trained with SFFA perform in general better than when learned via gradient back-propagation. An exception is Task IL, in which the layer-wise goodness function at the core of SFFA clashes with the need for sharing weights across tasks in this particular IL setting. This finding motivates future research paths aimed at overcoming this limitation with alternative formulations of goodness functions that account for the commonalities between tasks modeled incrementally by the CL model.

The rest of the paper is structured as follows: a brief review on CL and FFA is offered in Section \ref{sec:related_work}, whereas Section \ref{sec:methodology} delves into the proposed SFFA. Next, Section \ref{sec:increasing_pepe} elaborates on the motivation for using SFFA in CL scenarios, together with a reflection on the limitations posed by its adoption in this learning setting. Experiments are described and results are discussed in Sections \ref{sec:experimental_setup} and \ref{sec:results_discussion}, respectively. Finally, Section \ref{sec:conclusion} concludes the paper with a summary and prospects on future research lines stemming to this work.

\section{Related Work} \label{sec:related_work}

Before proceeding with the details of the proposed SFFA, we briefly revisit herein some background and relevant works in CL and FFA, leading to a short statement of the contribution of this manuscript within the reviewed literature.

\paragraph{Continual Learning} This learning paradigm is characterized by the continuous acquisition of knowledge needed to model non-stationary data streams. In this context, models trained within the CL framework are prone to suffering from catastrophic forgetting, where the assimilation of new concepts can be detrimental to previously gained knowledge \citep{goodfellow2013empirical}. Effectively navigating these scenarios requires a careful balance between plasticity, which facilitates the assimilation of novel concepts, and stability, crucial for retaining previously learned knowledge. 

While various scenarios fall within the scope of CL, this paper specifically concentrates on three categories: Class IL, Domain IL, and Task IL, which differ from each other depending on the type of new information to be incrementally incorporated to the CL model over time. These scenarios diverge from the conventional monolithic task definition adopted in non-CL problems, where datasets typically consist of a fixed set of data instances and label pairs. In contrast, CL datasets are composed of a collection of distinct tasks, each associated with different data instances and labels. Formally, this can be represented as $\{(X_i, Y_i)\}_{i\leq T}$, where $X_i$ denotes the data and $Y_i$ the labels of task $i$, and $T$ denotes the total number of tasks appearing sequentially over time. \textit{Class IL} occurs when the set of labels $Y_i$ expands along time, expressed as $Y_i \not \subseteq \cup_{j < i} Y_j$, whereas the data distribution of samples from previously encountered classes remains constant across all tasks. Although prior classes can reappear during training, it is a common practice to train models using disjoint sets of classes for each task, ensuring $Y_i \cap \cup_{j<i} Y_j = \emptyset$. Under this Class IL scenario, the models' architecture is designed so that the classification layer can be dynamically increased for the output layer to accommodate the total number of labels seen so far. This scenario is especially susceptible to the catastrophic forgetting effect due to the continued exposure of the model to instances belonging to previously characterized classes. On the other hand, \textit{Domain IL} involves scenarios where the distribution of tasks varies across tasks while the set of labels stays constant, i.e., $P(X_i) \neq P(X_j)$ for $j<i$, where $P(X)$ denotes the distribution of $X$. Such scenarios are common in non-stationary environments where models must classify objects whose domain expands over time. Lastly, \textit{Task IL} addresses scenarios where distinct problems are encoded as separate tasks, enabling the model to leverage task-specific information during its learning process. As data between these tasks is disjoint, most modeling approaches reported currently for Task IL rely on multi-head architectures, wherein each head is responsible for modeling one individual task. This architectural choice is shown to be effective in mitigating the overall impact of complementary tasks on each head. We refer to \citet{hsu2018re} for a more comprehensive understanding of the differences between these three CL scenarios.

From an algorithmic perspective, the review of CL methods contributed in \citet{de2021continual} suggested categorizing these techniques into three main groups: \textit{architecture-based methods}, which leverage the model to isolate weights for individual tasks; \textit{replay methods}, which reuse previously seen data to mitigate forgetting during the training of new tasks; and \textit{regularization methods}, which introduce regularization terms into the loss function to consolidate prior knowledge. Our work proposes a novel learning algorithm with various architectural modifications that will be later described in detail, hence we focus our literature review on the latter two categories. On one hand, replay methods have hitherto performed best compared to other algorithmic counterparts, albeit subject to higher memory requirements due to the need for storing a buffer of samples to be replayed \citep{yang2021benchmark}. These methods ensure optimal accuracy in scenarios where all previously seen data is stored in the replay buffer \citep{chaudhry2019tiny}. Another well-known replay technique is the Gradient Episodic Memory (GEM) algorithm, which utilizes a small memory to obtain the gradient projection direction from previous tasks, preventing the model's parameters from deviating in a direction contrary to old tasks \citep{lopez2017gradient}. In contrast, regularization methods do not impose memory constraints based on the data, resulting in more cost-effective algorithms at the expense of stability. The Elastic Weight Consolidation (EWC) algorithm regularizes the weights of the network using the diagonal of the Fisher information matrix of previous tasks as a measure of the weight importance \citep{kirkpatrick2017overcoming}. A similar approach is taken in Synaptic Intelligence (SI), where the computation of the weights' importance occurs in an online fashion, achieved by adding a surrogate loss to prevent the weights from diverging too far from previously learned weights \citep{zenke2017continual}. A similar effort is the Memory Aware Synapses (MAS) proposed by \citet{aljundi2018memory}, which is an online unsupervised technique that estimates the importance of weights by measuring to which extent small perturbations affect the network's weights.

In addition to the previously mentioned methods, an important branch of CL techniques closely related to the proposed algorithm are \textit{gradient-free} learning algorithms. Algorithms within this branch recall several inductive biases from the biological brain, such as sparsity and local plasticity, to avoid the catastrophic forgetting effect. One example of these algorithms was recently developed by \citet{lassig2023bio}, who proposed a biologically plausible hierarchical credit assignment model focusing on creating sparse representations within the latent space to improve the retention of knowledge during the different tasks. Similarly, \citet{ororbia2022lifelong} proposed a biologically inspired architecture with local synapse updates, achieving competitive accuracy levels when compared to back-propagation based approaches.
\vspace{-0.2cm}

\paragraph{The Forward-Forward Algorithm} Originally proposed by \citet{hinton2022forward}, FFA proposes a training procedure that is essentially distinct from the traditional gradient back-propagation algorithm. Specifically, FFA replaces the backward pass with a training loop composed of multiple forward passes, which minimize a set of local contrastive losses. This learning paradigm, introduced by \citet{kohan2018error,kohan2023signal}, belongs to the family of \textit{forward-only learning} algorithms, which eliminate the back-propagation step by resorting to a layer-wise training mechanism. One of the main motivations for these algorithms is to mitigate biological implausibilities introduced by the back-propagation algorithm. By adapting the credit assignment scheme to a layer-wise local optimization phase, these algorithms present further similarities to biological alternatives, closely resembling the behavior observed in biological brains \citep{zador2023catalyzing,ororbia2023brain}.

In each forward pass of FFA, the model evaluates the fitness of the input data for each layer. This fitness is denoted as the \textit{goodness} of a layer, and defined as $G: \mathbb{L}_\ell \rightarrow \mathbb{R}^+$, where $\mathbb{L}_\ell$ represents the latent space of layer $\ell$. During the initial forward pass, the model is fed with a sample from the training set $X$, referred to as a \textit{positive} sample. Conversely, in the second forward pass, the model processes an artificially generated sample from outside the data distribution of $X$, termed as a \textit{negative} sample. Throughout training, the model's parameters are learned to yield higher goodness scores on positive samples and low scores on negative samples. To achieve this, FFA trains each layer independently, computing its goodness and updating it based on the error relative to the expected goodness. The goodness function was originally defined as $G(\mathbf{l}_\ell) = \|\mathbf{l}_\ell\|^2$, where $\|\mathbf{l}_\ell\|$ denotes the Euclidean norm of the latent activations $\mathbf{l}_\ell$ of layer $\ell$. Under this definition, each layer maximizes its goodness score by maximizing its activity, while conversely achieving lower goodness scores over inactive layers.

As mentioned above, FFA enables layer-wise training, allowing each layer in the neural hierarchy to have its designated loss function and optimizer. In this regard, the Binary Cross-Entropy (BCE) loss is widely used due to the similarity of training a layer in FFA to a binary classification problem focused on detecting positive samples. To apply this loss function, Hinton defined the probability of a latent vector belonging to the positive set using a sigmoid function $\sigma(G(\mathbf{l}_\ell)-\theta)$, where $\mathbf{l}_\ell\in \mathbb{L}_\ell$ and $\theta$ is a threshold utilized to shift the probability function's range towards values close to zero. Within this framework, models can be trained layer-wise, utilizing the entire dataset for each layer individually, or in an online manner, training all layers with a batch before proceeding to the next batch. This diversity in the training pipeline offers additional solutions for systems with limited hardware capabilities, as models are not required to load all the weights during every step.

Since its inception, several works have proven the effectiveness of FFA as a practical alternative for modeling small-scale datasets, achieving competitive levels of accuracy when compared to networks trained with gradient back-propagation \citep{hinton2022forward,lorberbom2023layer}. On the negative side, loss functions used by FFA have been shown to exhibit an asymmetric behavior in updates between positive and negative samples, affecting the stability and convergence of the training process. To circumvent this issue, \citet{lee2023symba} proposed the use of an alternative loss function, replacing the BCE loss with a softplus function applied to the mean goodness difference between the positive and negative samples. This reformulation of the loss function results in an improved accuracy over the original FFA. In addition, the authors suggested an alternative method for generating negative samples, substituting the original one-hot label encoding with a random vector encoding. This change reduces the scalability restrictions posed by the original implementation of the algorithm. Further along this line, networks trained using FFA have been observed to develop sparse latent structures, showing a high neural specialization across distinct classes \citep{tosato2023emergent,yang2023theory}. This latent structure has also been identified by \citet{ororbia2023predictive}, who pointed at the representational capabilities of its latent space. This work showed that the network creates a latent structure containing sufficient information to reconstruct the original inputs with a simple decoder network.

A downside of the naive FFA is that fully-connected and locally-connected models trained in a layer-wise fashion are prone to overfitting, making this training algorithm unsuitable for learning models over complex data \citep{hinton2022forward}. To overcome this issue, \citet{scodellaro2023training} suggested using spatial patterns as an alternative to one-hot labels, enabling the model to uniformly embed the label pattern into the image. With this approach, they achieved an accuracy of 99.16\% on the MNIST dataset; however, no evidence was given for image classification datasets with non-uniform background. Another method was introduced by \citet{papachristodoulou2023convolutional}, which assigned specific classes to distinct groups of neurons aiming to maximize the competitiveness between groups by only maximizing the goodness of the group when a sample from the respective class is forwarded into the network. While these methods have demonstrated competitive performance over convolutional networks trained using back-propagation, they do not scale as effectively as the standard FFA for datasets with a large set of classes. 

\section{Proposed Symmetric Forward-Forward Algorithm (SFFA)} \label{sec:methodology}

As pointed by \citet{lee2023symba}, the original formulation of FFA exhibits an uneven loss surface, impacting the learning dynamics and leading to sub-optimal accuracy levels. This issue finds its roots on the asymmetrical definition of the probability function $p(x \text{ is positive})$ used to compute the likelihood of a data point being positive. In the original formulation, the probability is determined by the logistic function, which maps the domain $(-\infty, \infty)$ to $[0,1]$. However, as the original goodness function lies in the range $[0,\infty)$, it fails to cover the entire domain of the probability function. To mitigate this, the original definition subtracts a threshold $\theta$ from the goodness function $G(x)\equiv G(\mathbf{l}_\ell|\text{input $x$})$, thereby extending the reachable domain to include negative values. Despite this adjustment, the minimum value of $p(x \text{ is positive})$ never reaches zero\footnote{For instance, with the value $\theta=2$ used in the literature, the maximum value of $p(x \text{ is negative})$ is approximately $0.88$.}. Consequently, this limitation restricts the capability of the loss function to converge properly on negative samples, resulting in sub-optimal learning. Appendix \ref{ap:probability_function} provides a more in depth analysis on this problem.

\begin{figure}[t]
    \centering
    \includegraphics[width=1\textwidth]{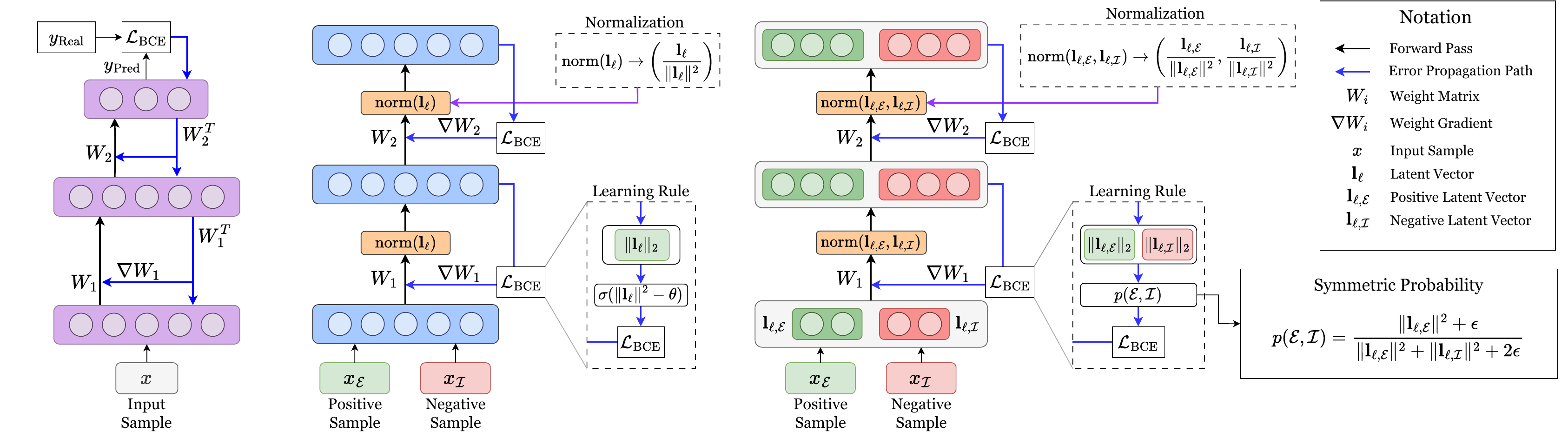}
    \resizebox{1\textwidth}{!}{\begin{tabular}{L{0.17\textwidth}L{0.3\textwidth}L{0.7\textwidth}}
         \hspace{1.2cm}(a) BP & \hspace{0.3cm}(b) FFA \citep{hinton2022forward} &  \hspace{1.1cm} (c) Proposed SFFA 
    \end{tabular}}
    \caption{Diagram illustrating the architecture of a) Backpropagation (BP), b) Forward-Forward Algorithm (FFA) \citep{hinton2022forward}, and c) Symmetric Forward-Forward Algorithm (SFFA). Each architecture highlights the input forward path (black arrows), and the error propagation path (blue arrows). Additionally, we describe the update mechanism of FFA and SFFA, and also show the probability function of SFFA.}
    \label{fig:models}
\end{figure}

To address this issue, in the \textit{Symmetric Forward-Forward Algorithm} proposed in this work, each layer is divided into positive and negative neurons, each set being specifically trained to attain high activity values when presented with samples of their respective types. More precisely, the positive set of neurons, denoted as $\mathcal{E}$, is designed to activate when processing positive samples, while the negative set of neurons, denoted as $\mathcal{I}$, is intended to achieve high activity values for negative samples. This division of neurons facilitates the formulation of a new goodness function, which can also serve as a probability function for the network. The goodness function $p : \mathbb{L}_\ell \times \mathbb{L}_\ell \rightarrow \mathbb{R}_{\geq 0}$, mapping two latent vectors to a non-negative real value, is defined as:
\begin{equation}
\label{eq:probability_function}
p(\mathcal{E}, \mathcal{I}) = \mathbb{P}\left(\{\mathbf{l}_{\ell,\mathcal{E}}, \mathbf{l}_{\ell,\mathcal{I}}\} \text{ are positive}\right) = \frac{\|\mathbf{l}_{\ell,\mathcal{E}}\|^2 + \epsilon}{\|\mathbf{l}_{\ell,\mathcal{E}}\|^2 + \|\mathbf{l}_{\ell,\mathcal{I}}\|^2 + 2\epsilon},
\end{equation}
where $\epsilon$ is a small value introduced to prevent division by zero. The positive latent vector $\mathbf{l}_{\ell,\mathcal{E}}$ and the negative latent vector $\mathbf{l}_{\ell,\mathcal{I}}$ work together to create a balanced representation, ensuring a symmetric negative probability function $p(\mathcal{E}, \mathcal{I}) = 1 - p(\mathcal{I}, \mathcal{E})$.

As the function $p\left(\mathcal{E}, \mathcal{I}\right)$ computes the ratio between the activities of the two sets, distinct training processes emerge. While the FFA maximizes the goodness score by increasing the activity of each layer when exposed to positive samples, our approach achieves higher goodness values by elevating the ratio between the activities of positive and negative neurons. By means of this strategy, our proposed method efficiently computes the goodness score even when utilizing layer normalization. This feature is particularly interesting when working with bounded activation functions, as they generate less sparse latent structures compared to their ReLU-like counterparts. Furthermore, to improve the sparsity of the latent output, we advocate for the integration of lateral inhibition mechanisms. One pioneering approach was originally introduced by \citet{ororbia2023predictive}, where they incorporated a learnable recurrent set of weights at each layer. This recurrent dynamic allowed the network to exhibit inhibitory and excitatory self-regulation. This form of inhibition is particularly suitable when using the original \textit{static video} approach for image data \citep{hinton2022forward}, where a single forward pass is replaced by a recurrent time-dependent pass. In our work, since networks only employ a single forward pass per input, we utilize a k-WTA (\emph{Winner Takes All}) scheme, which sets to zero all neurons in each layer, except the $k$ most active ones.

As the goodness information of an input is represented by the activity of the latent vector of each layer, a normalization step was introduced in FFA to prevent subsequent layers from using this information for the prediction. Since the original FFA only employs the norm of the vector to compute the probability, the normalization consisted of mapping the latent vector to its normalized version, i.e., $\mathbf{l}_{\ell} \rightarrow \mathbf{l}_{\ell}/\|\mathbf{l}_{\ell}\|$. This results in all samples having the same goodness, thereby unbiasing the latent vector for the next layers. To mimic this process in SFFA, we require a normalization process in which every latent vector achieves the same goodness score regardless of their  value. Given the probabilistic nature of our defined goodness function, we target a final goodness score equal to $0.5$, which fixes the probability of latent vectors into the middlepoint between positive and negative scores, thereby unbiasing them for subsequent layers. Since our function measures the ratio between positive and negative samples, to achieve the desired score, the squared norm of positive and negative latent should be equal. Given these properties, we define our normalization function as $\texttt{norm} : \mathbb{L_{\ell}} \times \mathbb{L_{\ell}} \rightarrow \mathbb{L_{\ell}} \times \mathbb{L_{\ell}}$ to normalize each neural set independently to the same value. This normalization is given by:
\begin{equation}
\texttt{norm}(\mathbf{l}_{\ell}) = \texttt{norm}(\{\mathbf{l}_{\ell,\mathcal{E}}, \mathbf{l}_{\ell,\mathcal{I}}\}) = \left(\frac{\mathbf{l}_{\ell,\mathcal{E}}}{\|\mathbf{l}_{\ell,\mathcal{E}}\|^2}, \frac{\mathbf{l}_{\ell,\mathcal{I}}}{\|\mathbf{l}_{\ell,\mathcal{I}}\|^2} \right),
\end{equation}
where $\mathbf{l}_{\ell}$ is the latent vector, and $\mathbf{l}_{\ell,\mathcal{E}}$ and $\mathbf{l}_{\ell_\mathcal{I}}$ are the latent vectors corresponding to the positive and negative samples, respectively.

\section{SFFA for Continual Learning}
\label{sec:increasing_pepe}

This section outlines the motivation for incorporating Forward-Forward-like algorithms into CL pipelines. Furthermore, we introduce a method for adapting these algorithms to both Class and Task IL. Lastly, we address potential limitations that may arise from the default formulation of the algorithm.

\paragraph{Motivation} Examining the potential of FFA-like methods for CL resides mainly in their representational properties. Specifically, the generated latent spaces possess a sparse structure characterized by neurons with a high degree of class specialization (see Appendix \ref{sec:sparsity_jiji} for a detailed analysis of the sparsity induced by FFA and the proposed SFFA). We hypothesize that these features may help in reducing the number of updates on trained neurons, giving preference to the least active or less specialized neurons for each task. Consequently, this property could help mitigate the impact of the catastrophic forgetting effect. A second motivation stems from the layer-wise training mechanism employed by the FFA, facilitating training for low-resource systems and enabling seamlessly scalable systems. For instance, since each layer is trained independently, increasing the number of layers can be achieved without requiring additional training for previous layers. This process seamlessly adapts to the previous output latent space, notably reducing the total computing capacity required. Furthermore, training models with this algorithm eliminates the need for maintaining the entire model in memory, thereby reducing significantly the required memory allocated for the model.

\paragraph{Adaptation} Different CL scenarios often require architectural modifications to prevent the model from forgetting previously acquired knowledge. For instance, while Domain IL models can reuse conventional classifier architectures, Class IL scenarios require expanding their number of outputs as the total number of classes grows over time. Similarly, Task IL commonly employs multi-head approaches, where each classifier is only active based on the given label. In contrast, networks trained by using the FFA do not contain a classification layer and, therefore, cannot employ the aforementioned architectural modifications. Similar to conventional models, Domain IL scenarios do not require any architectural change, as they maintain a fixed label set over different tasks. However, Class IL and Task IL scenarios have to adapt their label encoding techniques to operate through the growing set of labels and tasks. 

To adapt the models to these paradigms, we propose to modify the negative forward phase. The standard approach typically involves forwarding one positive sample and one negative sample, usually created by selecting an instance from the training dataset and embedding an incorrect label in it. In our case, we employ all labels one by one, storing the goodness of each in a vector. By using each coordinate of the vector independently, we can train the model using the standard BCE loss. In the case of Class IL, whenever a new label is encountered, it should be registered in a list of known classes. This allows the goodness vector to increase in size as new labels arrive over time. In contrast, for Task IL, we propose a method in which we only forward the labels of the respective task. We omit forwarding labels from other tasks, as within the respective task space, they would be neither negative nor positive. The algorithm employed to adapt FFA and SFFA to CL scenarios are detailed in Algorithms \ref{algo:cil_set} and \ref{algo:til_set}, present in Appendix \ref{ap:adaptation_algorithms}.

\paragraph{Limitations} A shortcoming of SFFA emerges from the prevalence of multi-head layers in neural networks used for Task IL scenarios. These layers play a crucial role in isolating task-specific weights during the training phase, thereby mitigating the impact of catastrophic forgetting in scenarios with increasing number of tasks. Given the layer-wise training approach of FFA-like approaches, all layers can be simultaneously regarded as feature extractors and classification layers. To effectively implement this layer-wise strategy, it becomes necessary to isolate task-dependent weights from each task, which results in a set of disjoint models. This contradicts a fundamental aspect of CL, by which models are expected to share knowledge from previous tasks to facilitate the on-line learning of new tasks. Another significant limitation is the absence of CL techniques specifically tailored for models trained using FFA-based algorithms. Most existing techniques are developed for models trained using back-propagation, and seamlessly adapting them into our herein proposed SFFA is not straightforward. For instance, algorithms that rely on information produced in the last non-classification layer cannot be applied, since in SFFA all layers inherently serve as classification layers. This limitation is particularly evident in the case of the \textit{Less-forgetful Learning} algorithm \citep{jung2016less}, which operates on the feature space towards reducing the impact of catastrophic forgetting. As later outlined in the manuscript, future research will target these and other limitations, departing from the promising performance shown by SFFA in the experiments next presented.

\section{Experimental Setup}
\label{sec:experimental_setup}

To thoroughly assess the performance of SFFA and its adaptability to CL scenarios, we formulate two research questions that will be addressed through experiments and a further analysis of its results:
\begin{itemize}[leftmargin=*]
    \item RQ1: Does the removal of the asymmetry in the SFFA goodness function enhance its generalization capabilities?
    \item RQ2: Can the proposed SFFA be used with conventional CL methods to solve IL scenarios?
\end{itemize}

To address RQ1, we conduct experiments to comprehensively compare the performance of a model trained via SFFA to the same model trained with the original FFA and a naive gradient back-propagation algorithm. Each experiment involves training a neural network using one of the three training mechanisms. To ensure a fair comparison between the trained models, they are configured with an identical number of neurons, hidden layers, and learning rate. Each training run consists of 100 epochs to guarantee convergence. The comparison is performed by measuring the maximum accuracy achieved on the test dataset during the training phase. Detailed information about the hyper-parameters used in these experiments are given in Appendix \ref{ap:hyperparameters}.

To ensure a diverse range of data distributions, four distinct image classification datasets are selected. The first three datasets are MNIST, Fashion MNIST (FMNIST) and KMNIST, each containing 60,000 training images and 10,000 test images evenly distributed across 10 classes. Additionally, we utilize the EMNIST Letters dataset, consisting of 145,000 images distributed across 26 labels. All datasets consist of gray-scale images with dimensions $28\times 28$ pixels.  A preprocessing step is applied to normalize all images. No data augmentation pipeline is performed. Nevertheless, due to the current lack of robust convolutional adaptations of FFA-like methods, the selected datasets have been restricted to those with uniform backgrounds. As detailed developed in the Future Work (Section \ref{sec:conclusion}), addressing this limitation is a priority for future papers.

Another crucial step in our experiments is the inclusion of a regularization term into the loss function. This is motivated by the unstable behavior of the gradient under vectors with low activity. Appendix \ref{ap:gradient} offers an in-depth discussion about this noted instability. Given that our algorithm lacks control over layer activity, models may undergo changes over time, potentially leading to near-vanishing activity, which can produce exploding dynamics in the gradient. To counteract this effect, we introduce an additional multiplicative Gaussian term into the loss function. This term actively pushes the latent space away from zero, preventing the model from approaching low-activity regions. Consequently, the loss function is modified as:
\begin{equation}
\label{eq:loss_regu}
    \mathcal{L}(\mathbf{L}_{\ell}) = \left(1+e^{-\alpha\norm{\mathbf{L}_{\ell}}_1}\right)  \mathcal{L}_{\textup{BCE}}(\mathbf{L}_{\ell}),
\end{equation}
where $\mathbf{L}_{\ell} = \{\mathbf{l}_{\ell, 1}, \mathbf{l}_{\ell, 2},\ldots\}$ represents a batch of latent vectors of layer $\ell$, and $\alpha$ is a scaling hyper-parameter that measures the pushing distance and force.

In addition to the prior evaluation of SFFA in static environments, our goal in RQ2 is to analyze the adaptability of our approach within CL environments. To achieve this, we conduct multiple experiments comparing the results corresponding to the network trained by using the back-propagation algorithm with those of the model learned by using SFFA. Each experiment involves a series of disjoint image classification tasks where the model has to learn novel classes while retaining knowledge of previously learned data. Furthermore, each experiment utilizes a CL technique from the literature, allowing us to gauge the adaptability of both training mechanisms to each considered CL approach. To measure the models' accuracy, we compute the classification accuracy of each experiment over the joint test dataset, averaging the results over 10 different runs to account for the statistical significance of the performance gaps observed between the comparison counterparts.

The process used to adapt SFFA for CL scenarios is outlined in Algorithms \ref{algo:cil_set} (Class IL) and \ref{algo:til_set} (Task IL). Due to its fixed label set, Domain IL does not require additional modifications. Given that CL scenarios demand greater stability and lower processing latency, we adjust the number of epochs for each task training step. Specifically, the epochs were reduced from $100$ in RQ1 to $2$. All hyper-parameter values and metrics used in these experiments are detailed in Appendix \ref{ap:hyperparameters}.

The five CL techniques employed in our experiments are \emph{Elastic Weight Consolidation} (EWC \citep{kirkpatrick2017overcoming}), \emph{Synaptic Intelligence} (SI \citep{zenke2017continual}), \emph{Random Replay} (REP \citep{yang2021benchmark}), \emph{Gradient Episodic Memory} (GEM \citep{lopez2017gradient}), and \emph{Memory Aware Synapses} (MAS \citep{aljundi2018memory}). We choose these specific algorithms to cover both regularization and replay methods, intentionally avoiding architecture-based approaches. This decision is driven by the aforementioned incompatibility between architecture-based techniques designed for back-propagation (e.g. multiple task-specific heads) and our novel model design for SFFA. To ensure a smooth integration of SFFA into these CL frameworks, we implement our algorithm within the Avalanche ecosystem \citep{lomonaco2021avalanche}. This integration not only facilitates the incorporation of Avalanche's CL plugins, but also ensures a cohesive interaction with models trained using SSFA. This approach also aims to enhance the compatibility and interoperability of SFFA within the broader CL research community, stimulating follow-up studies that consider this proposed training algorithm. The experiments for RQ2 consider two datasets: SplitMNIST and SplitFMNIST. These datasets are obtained by randomly partitioning the original MNIST and FMNIST datasets into several disjoint sets, ensuring that each class appears in only one set. Specifically, the total number of classes is divided into 5 tasks, with each task containing 2 classes.

\section{Results and Discussions}
\label{sec:results_discussion}

This section presents the results obtained from the experiments, coupled with a discussion that addresses the research questions posed earlier: RQ1 (Section \ref{ssec:rq1}) and RQ2 (Section \ref{ssec:rq2}). The analysis delves into both the effectiveness and weaknesses of the proposed SFFA.

\subsection{RQ1: Does the removal of the asymmetry in the SFFA goodness function enhance its generalization capabilities?} \label{ssec:rq1}

The results obtained for RQ1 presented in Table \ref{tab:main_accuracy} underscore the superior generalization capabilities of the model trained with SFFA when compared to the same model learned via the original FFA. Moreover, SFFA exhibits competitive performance in comparison to the model using gradient back-propagation on the MNIST, FMNIST, and KMNIST datasets, with only marginal accuracy losses. Notably, SFFA achieves comparable accuracy levels in FMNIST, substantially reducing the accuracy gap of FFA from $3.72$ to a negligible $0.08$. When focusing on the EMNIST dataset, despite a wider accuracy drop of over $3$ points for both FFA and SFFA, SFFA still performs better than FFA by $1.38$ points. These results are consistent with our initial hypothesis, which posits that by eliminating the inherent asymmetry from the algorithm, the model's accuracy is improved.
\begin{table}[h]
    \centering
    \caption{Best classification accuracy attained during 100 epochs for the different training algorithms on MNIST, Fashion MNIST (FMNIST), Kanji MNIST (KMNIST) and EMNIST Letters (EMNIST) datasets. Gaps with respect to gradient back-propagation are indicated in red. The accuracies are measured based on non-CL metrics.}
    \resizebox{0.7\textwidth}{!}{\begin{tabular}{lllll}
        \toprule
      & MNIST & FMNIST & KMNIST & EMNIST \\
      \midrule
      Gradient back-propagation & 98.25 & 89.47 & 92.11 & 92.19 \\
      FFA             & 97.77 \accdiff{0.48}  & 85.75 \accdiff{3.72} & 91.86 \accdiffb{0.25} & 87.25 \accdiff{4.94} \\
      SFFA (proposed)   & 97.87 \accdiffb{0.38}  & 89.39 \accdiffb{0.08} & 90.47 \accdiff{1.64} & 88.63 \accdiffb{3.56} \\
      \bottomrule
    \end{tabular}}
    \label{tab:main_accuracy}
\end{table}

\subsection{RQ2: Can SFFA be used with conventional CL methods to solve IL scenarios?} \label{ssec:rq2}

\begin{figure}[!b]
\centering
    \includegraphics[width=0.9\textwidth]{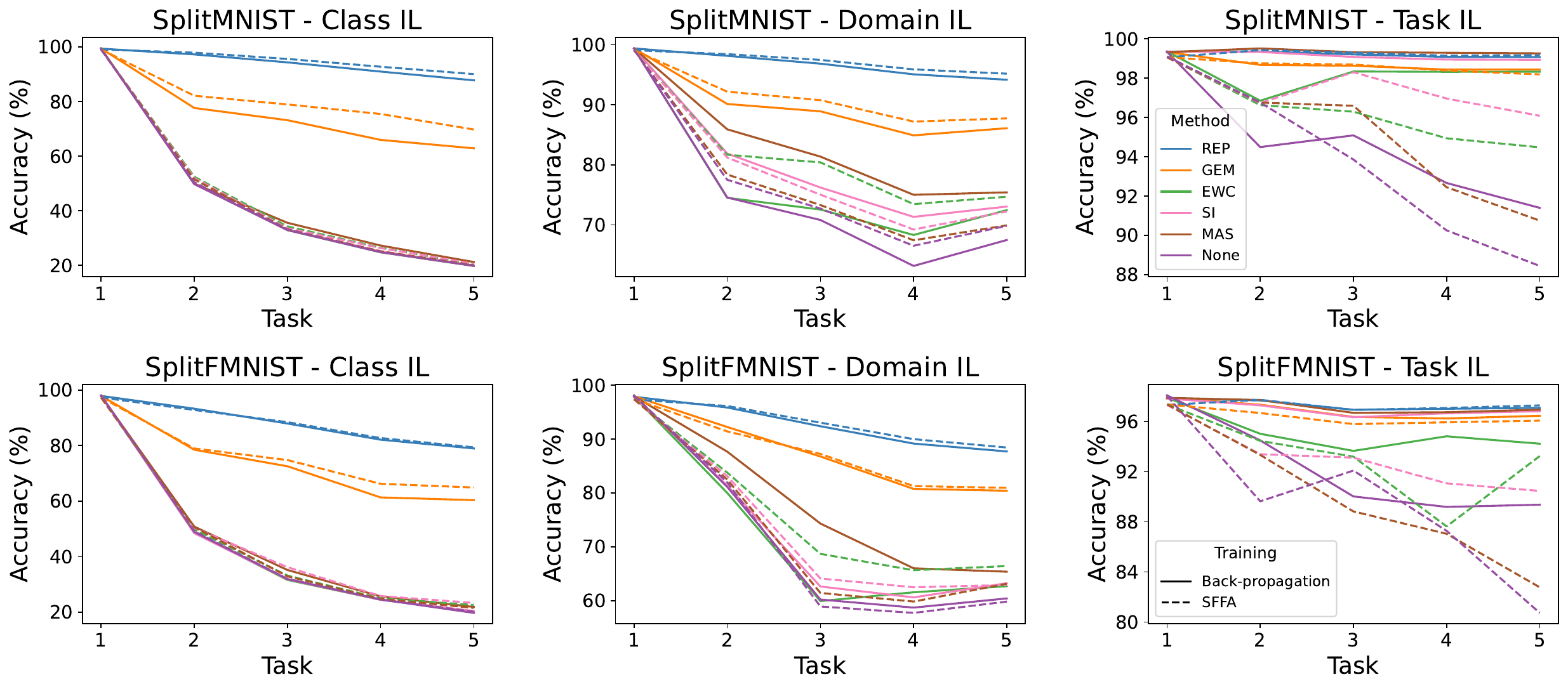}
    \caption{Average classification accuracy of the gradient back-propagation algorithm and SFFA for different IL scenarios: Class IL (first column), Domain IL (center column) and Task IL (right column). Each point captures the average accuracy of the model (over 10 runs) on all evaluation data from tasks that the model has been previously trained on.}
    \label{fig:result_acc_on_task}
\end{figure}

We now shift the focus of the discussion towards RQ2, pausing first at the accuracy of the experiments conducted on the SplitMNIST and SplitFMNIST datasets shown in Table \ref{tab:mnist_cl_results}, and $Fgt$ and $FwT$ metrics shown in Table \ref{tab:mnist_fgt_fwt_results} in Appendix \ref{ap:additional_results_rq2}. Furthermore, Figure \ref{fig:result_acc_on_task} depicts the models' performance across the different tasks for each experiment. These plots allow assessing the relative accuracy performance achieved on a unique task after learning a varying number of previous tasks.

The Class IL scenario proves to be the most challenging one among the three IL setups under study. Indeed, several techniques struggle to surpass the accuracy achieved by a random classifier. This shortcoming is reflected in the performance of EWC, SI, and MAS, all achieving accuracy levels hovering around the 20\% mark on both datasets. This outcome proves that regularization techniques underperform when retaining knowledge, which was previously observed by \citet{hsu2018re} with models trained using gradient back-propagation. In this case, FFA is the only algorithm capable of attaining a better than random accuracy when employing EWC and SI. However, the high variance in the results is also remarkable. These results arise as the algorithm is able to retain knowledge in half of the trials, while underperforming in the other half. In contrast, replay-based techniques consistently exhibit superior performance across both datasets, with random replay emerging as the top-performing algorithm. Notably, these techniques yield comparable results when compared to the back-propagation algorithm and SFFA. However, it is important to highlight the slight advantage obtained when employing the GEM algorithm, achieving a $7$ point and $4$ point accuracy improvement in MNIST and FMNIST when using SFFA instead of back-propagation, and the $12$ and $8$ points of accuracy gained when using FFA against back-propagation, respectively.

Results in the Domain IL scenario closely resemble those in Class IL, with replay-based algorithms consistently achieving higher accuracies, and random replay emerging as the best performing algorithm in the benchmark. Similar to the previous scenario, FFA and SFFA achieve slightly higher accuracies in replay-based methods compared to the backpropagation algorithm, although the difference is narrower than in Class IL. Differently from our previous discussion, additional CL techniques contribute to improved accuracies, where networks trained using only gradient back-propagation exhibit lower accuracy than those trained using EWC or SI. Similarly to the previous case, FFA is more suitable for these algorithms than SFFA, but still feature a higher variability in its accuracy scores. Remarkably, MAS produces varying results depending on the training algorithm used, achieving less accurate results in FFA-like models than in their back-propagation based counterparts. Additionally, the results in Table \ref{tab:mnist_cl_results} reveal a lower standard error of the mean in replay methods, indicating the higher robustness obtained by these models. Importantly, the standard error of the mean remains consistent across both training algorithms, evincing that SFFA does not yield a higher variability in the accuracy of the trained model. FFA remains as the most variable algorithm of the three approaches under comparison.
\begin{table}[h]
    \centering
    \caption{Averaged classification accuracy and standard error of the mean obtained over 10 experiments with different seeds for the Split MNIST and Split Fashion-MNIST datasets. The highest accuracy among all the methods has been marked in bold at for each dataset and CL scenario.}
    \resizebox{\textwidth}{!}{\begin{tabular}{llccccccccccc}
        \toprule
     \multicolumn{13}{c}{\multirow{2}{*}{Split MNIST}}\\
            \\ \midrule
        && \multicolumn{3}{c}{Class IL} && \multicolumn{3}{c}{Domain IL} && \multicolumn{3}{c}{Task IL}\\
                \cmidrule{3-5} \cmidrule{7-9} \cmidrule{11-13}
        Method & & Back-propagation & FFA & SFFA && Back-propagation & FFA & SFFA && Back-propagation & FFA &SFFA\\
        \midrule
        Adam    && \(19.81 \pm 0.04\) & \(18.34 \pm 2.72 \) & \(19.85 \pm 0.03\) && \(67.50 \pm 2.32\) & \(70.87 \pm 7.08 \) & \(69.86 \pm 2.13\) && \(91.40 \pm 1.34\) & \(90.01 \pm 3.66 \) & \(88.44 \pm 1.40\) \\
        
        EWC     && \(19.77 \pm 0.04\) & \(26.79 \pm 8.28 \) & \(20.32 \pm 0.22\) && \(72.47 \pm 2.35\) & \(78.54 \pm 6.50 \) & \(74.70 \pm 2.31\) && \(98.33 \pm 0.17\) & \(96.55 \pm 1.24 \) & \(94.48 \pm 1.01\) \\
        SI      && \(19.81 \pm 0.03\) & \(28.01 \pm 7.77 \) & \(20.50 \pm 0.19\) && \(73.06 \pm 2.54\) & \(77.32 \pm 5.98 \) & \(72.25 \pm 1.93\) && \(98.93 \pm 0.10\) & \(96.43 \pm 2.42 \) & \(96.08 \pm 0.62\) \\
        MAS     && \(21.21 \pm 0.54\) & \(19.18 \pm 5.78 \) & \(20.04 \pm 0.11\) && \(75.42 \pm 1.80\) & \(71.04 \pm 7.03 \) & \(69.93 \pm 2.09\) && \(\textbf{99.26} \pm \textbf{0.05}\) & \(90.11 \pm 4.07 \) & \(90.75 \pm 1.03\) \\
        Replay  && \(87.75 \pm 0.26\) & \(81.28 \pm 2.74 \) & \(\textbf{90.03} \pm \textbf{0.24}\) && \(94.16 \pm 0.50\) & \(93.01 \pm 1.64 \) & \(\textbf{95.17} \pm \textbf{0.35}\) && \(99.07 \pm 0.05\) & \(98.34 \pm 0.33 \) & \(99.17 \pm 0.09\) \\
        GEM     && \(62.85 \pm 0.84\) & \(74.62 \pm 3.87 \) & \(69.73 \pm 0.57\) && \(86.10 \pm 1.02\) & \(88.26 \pm 2.88 \) & \(87.73 \pm 0.86\) && \(98.44 \pm 0.17\) & \(97.15 \pm 0.79 \) & \(98.19 \pm 0.25\) \\
     \midrule
     \multicolumn{13}{c}{\multirow{2}{*}{Split Fashion-MNIST}}\\
            \\
     \midrule 
        && \multicolumn{3}{c}{Class IL} && \multicolumn{3}{c}{Domain IL} && \multicolumn{3}{c}{Task IL}\\
        \cmidrule{3-5} \cmidrule{7-9} \cmidrule{11-13}
        Method & & Back-propagation & FFA & SFFA && Back-propagation & FFA & SFFA && Back-propagation & FFA & SFFA\\
        \midrule
        Adam    && \(19.74 \pm 0.07\)& \(21.98 \pm 3.11 \) & \(20.18 \pm 0.25\)&& \(60.42 \pm 2.50\)& \(64.25 \pm 9.72 \) & \(59.84 \pm 2.75\)&& \(89.37 \pm 1.37\)& \(78.16 \pm 10.73 \) &\(80.74 \pm 2.41\)\\
        
        EWC     && \(20.21 \pm 0.48\)& \(23.54 \pm 5.44 \) & \(22.58 \pm 1.20\)&& \(62.69 \pm 2.77\)& \(69.03 \pm 8.65 \) & \(66.43 \pm 2.52\)&& \(94.23 \pm 1.42\)& \(93.14 \pm 2.73 \) & \(93.25 \pm 1.51\)\\
        SI      && \(20.41 \pm 0.59\)& \(25.84 \pm 4.05 \) & \(23.33 \pm 1.04\)&& \(63.24 \pm 2.55\)& \(68.59 \pm 8.40 \) & \(62.91 \pm 2.58\)&& \(96.82 \pm 0.59\)& \(90.59 \pm 7.00 \) & \(90.47 \pm 1.52\)\\
        MAS     && \(21.88 \pm 0.83\)& \(22.17 \pm 5.61 \) & \(21.54 \pm 0.80\)&& \(65.38 \pm 2.65\)& \(64.48 \pm 9.61 \) & \(63.15 \pm 3.14\)&& \(96.95 \pm 0.55\)& \(79.97 \pm 7.83 \) & \(82.78 \pm 2.52\)\\
        Replay  && \(78.97 \pm 0.26\)& \(72.85 \pm 3.62 \) & \(\textbf{79.38} \pm \textbf{0.27}\)&& \(87.69 \pm 0.47\)& \(86.58 \pm 1.87 \) & \(\textbf{88.42} \pm \textbf{0.47}\)&& \(97.09 \pm 0.45\)& \(96.83 \pm 1.54 \) & \(\textbf{97.29} \pm \textbf{0.46}\)\\
        GEM     && \(60.35 \pm 1.33\)& \(68.07 \pm 1.42 \) & \(64.90 \pm 0.93\)&& \(80.40 \pm 1.25\)& \(82.19 \pm 2.56 \) & \(80.93 \pm 0.95\)&& \(96.46 \pm 0.46\)& \(95.62 \pm 2.21 \) & \(96.07 \pm 0.68\)\\
     \bottomrule
     
    \end{tabular}}    
    \label{tab:mnist_cl_results}
\end{table}

Lastly, in the Task IL scenario, the obtained results indicate that models trained with back-propagation consistently outperforms FFA-like algorithms in all CL methods, except for replay-based techniques. This discrepancy is clearly observed in the performance difference between models employing EWC and SI. In the case of EWC, there is an accuracy drop from $98.33$ in back-propagation to the result of FFA-like of $96.55/94.48$ in MNIST, and from $94.23$ to $93.14/93.25$ in FMNIST. A more significant difference is observed in SI, with a drop from $98.93$ to $96.43/96.08$ and from $96.82$ to $90.59/90.47$, respectively. Furthermore, MAS also demonstrates suboptimal performance, reinforcing the observed trend in the Domain IL scenario, where it works more effectively with backpropagation than with FFA or SFFA. However, this phenomenon does not appear in replay-based techniques. As highlighted in Section \ref{sec:increasing_pepe}, the limitation of non-separability in the layers of the model trained using FFA-like algorithms may reduce the accuracy in Task IL settings due to the need for weight sharing across tasks, resulting in the lower accuracy levels observed between back-propagation and FFA/SFFA in these Task IL experiments.

Over all experiments, the random replay technique exhibits a clear dominant trend, yielding the highest accuracy and robustness across all scenarios. We posit that the decline in results when no replay is employed stems from the imbalanced data fed into the network. Once the model is trained on a specific class, no additional instances of that class are shown, resulting in a lack of positive activation. However, during the training of further classes, all existing labels will be forwarded into the model, expecting negative activity over previously seen classes, hence causing the model to increasingly respond negatively to previously seen labels. Nevertheless, as provided by the results in Domain IL and Task IL, currently existing techniques can be used in order to mitigate the catastrophic forgetting effect in these networks. This is the case of EWC and SI, which do not require additional adaption and achieve competitive accuracy on both datasets. Across all experiments, FFA and SFFA have attained similar forgetting and forward-transfer scores to back-propagation based networks, with FFA achieving even lower forgetting values in CI Tasks. Additionally, our SFFA model exhibits a more stable and controllable performance, showcasing a reduced variance within the performed CL experiments. While FFA may show increased accuracy in short-term experiments, the experienced variance can lead to suboptimal behavior on the long term, hindering its capability to serve as a optimal CL model. The positive results of these experiments clearly indicate the effectiveness of integrating conventional CL techniques with the proposed SFFA, concluding with a positive answer to the second research question RQ2.

\section{Conclusions and Future Work}
\label{sec:conclusion}

This study has introduced the Symmetric Forward-Forward Algorithm (SFFA), a novel formulation of the FFA designed to address the inherent asymmetry in the loss function. Through a series of image classification experiments, we have provided informed insights into the accuracy gains achieved by our approach. Additionally, considering the representational capabilities and sparse activations produced by SFFA, we have explored the potential of Forward-Forward-like algorithms for CL scenarios as an alternative to gradient back-propagation algorithms. The experimental results reported in this manuscript have illustrated that the SFFA can be seamlessly integrated with conventional CL techniques, effectively mitigating the impact of catastrophic forgetting in incremental learning tasks. 

Several conclusions have been drawn. To begin with, not all regularization techniques for CL can be trained with forward-forward algorithms, since this training algorithm does not decouple feature extraction from classification throughout the neural hierarchy. Furthermore, some algorithms do not perform well alongside FFA or SFFA, as observed in the case of the MAS algorithm. Within the usable algorithms, they exhibit similar levels of accuracy across both Class IL and Domain IL when compared to back-propagation, with only a slight performance gap observed for Task IL scenarios. This discrepancy is attributed to architectural differences between the networks, as current Forward-Forward-like networks lack a multi-head layer, thereby preventing the parameter isolation offered by back-propagation. Above all, the results presented in this paper underscore a promising future for the application of Forward-Forward-like algorithms to CL tasks.

Several research directions have been identified rooted on the overall good results of this study. To begin with, we plan to narrow the performance gap observed for SFFA in Task IL scenarios by incorporating architectural modifications to the neural model that allow for the use of task-specific parameter isolation strategies. Furthermore, based on the sparsity analysis made in our experiments (see Appendix \ref{sec:sparsity_jiji}), techniques based on parameter isolation/masking \citep{konishi2023parameter} could reduce the catastrophic forgetting effect in Class IL scenarios. In addition, due to their similarities with the FFA behavior, we aim at adapting techniques based on orthogonal projections \citep{lin2022beyond,guo2022adaptive}, as they could potentially improve the performance in Class IL scenarios without requiring memory-based schemes. Moreover, we plan to address the current limitations observed when working with complex image classification datasets, where the lack of scalable convolutional architectures hinders the practical usage of forward-only learning algorithms. For this purpose, we hypothesize that the development of tailored negative data generators based on input data can override the need for wave-like label embedding and can allow for the incorporation of convolutional and pooling layers in the neural architecture.

\section*{Acknowledgements} The authors thank the Basque Government for its funding support via the consolidated research groups MATHMODE (ref. T1256-22) and D4K (ref. IT1528-22), and the BEREZ-IA ELKARTEK project (ref. KK-2023/00012).

\bibliography{collas2024_conference}
\bibliographystyle{collas2024_conference}

\appendix

\section{Algorithm to Adapt FFA to CL}
\label{ap:adaptation_algorithms}

In this appendix we present the algorithms used to adapt FFA-like algorithms to CL scenarios. Algorithm \ref{algo:cil_set} modifies the negative generation of CIL scenarios, so that only previously seen classes are embedded as negative data during each task. Similarly, Algorithm \ref{algo:til_set} describes the adaptation for TIL tasks, ensuring that only samples from each task are used during negative data creation, and that during prediction, only samples with labels from the tasks are compared.
\begin{figure}[h]
\begin{minipage}{0.48\linewidth}
\begin{algorithm}[H]
\label{algo:cil_set}
\SetAlgoNlRelativeSize{0}
\SetAlgoNlRelativeSize{-1}
\DontPrintSemicolon
\caption{FFA/SFFA adapted to Class IL}
\KwData{Model $\mathcal{N}$, list of encoding vectors $\mathcal{V}$ and set of seen classes $\mathcal{C}$}

\SetKwFunction{FMain}{Adapt}
\SetKwProg{Fn}{Function}{:}{}
\Fn{\FMain{\textup{Task} $T$}}{
    Let $C$ be the classes of task $T$\;
    \For{\textup{class} $c$ \textup{in} $C$}{
        \If{$c \not \in \mathcal{C}$}{
            Create encoding pattern for $c$ in $V$\;
            Append class $c$ to set $\mathcal{C}$\;
        }
    }
}
  
\SetKwFunction{FMainn}{Forward}
\Fn{\FMainn{\textup{batch} $X$}}{
    Create goodness vector $\mathbf{g} \gets \mathbf{0}_{|\mathcal{C}|}$\;
    \For{\textup{class} $c$ \textup{in} $\mathcal{C}$}{
        \For{\textup{each layer} $\ell$}{
            Obtain vector $\mathbf{l}_{\ell}$ from $X$ using $c$ \;
            $\mathbf{g}_c \gets $ goodness of $\mathbf{l}_{\ell}$ \;
        }
    }

    \KwRet $g$\;
}

\end{algorithm}
\end{minipage}\hspace{0.02\linewidth}
\begin{minipage}{0.48\linewidth}
\begin{algorithm}[H]
\DontPrintSemicolon
\label{algo:til_set}
\SetAlgoNlRelativeSize{0}
\SetAlgoNlRelativeSize{-1}
\caption{FFA/SFFA adapted to Task IL}
\KwData{Model $\mathcal{N}$, list of encoding vectors $\mathcal{V}$ and map of classes per task $\mathcal{C}$}
\SetKwFunction{FMain}{Adapt}
\SetKwProg{Fn}{Function}{:}{}
\Fn{\FMain{\textup{task} $T$, \textup{task index} $i$}}{
    Let $C$ be the classes of task $T$\;
    \For{\textup{class} $c$ \textup{in} $C$}{
        \If{$c \not \in \mathcal{C}_i$}{
            Create encoding pattern for $c$ in $V$\;
            Append class $c$ to set $\mathcal{C}_i$
        }
    }
}
\SetKwFunction{FMainn}{Forward}
\Fn{\FMainn{\textup{batch} $X$, \textup{task Label} $i$}}{
    Create goodness vector $\mathbf{g} \gets \mathbf{0}_{|\mathcal{C}_i|}$\;
    \For{\textup{class} $c$ \textup{in} $\mathcal{C}_i$}{
        \For{\textup{each layer} $\ell$}{
            Obtain vector $\mathbf{l}_{\ell}$ from $X$ using $c$ \;
            $\mathbf{g}_c \gets $ goodness of $\mathbf{l}_{\ell}$ 
        }
    }

    \KwRet $g$\;
}

\end{algorithm}
\end{minipage}

\end{figure}

\section{Comparison of the Probability Functions of FFA and SFFA} \label{ap:probability_function}

\begin{figure}[h]
    \centering
    \includegraphics[width=0.8\textwidth]{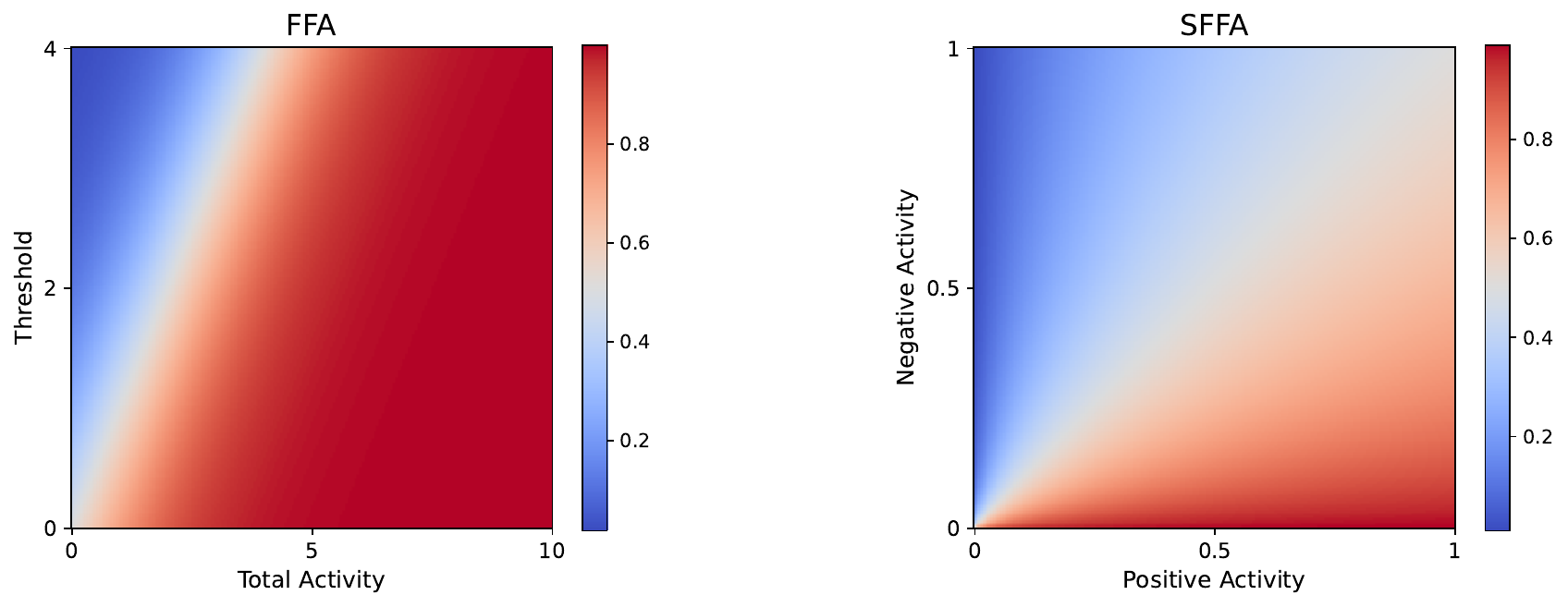}
    \caption{Visualization of the probability function of FFA (left) and the SFFA (right). The x-axis in the plot corresponding to FFA contains the total activity of the layer, whereas its y-axis specifies the threshold. The x-axis of the SFFA plot denotes the activity of the positive neurons, whereas the y-axis depicts the activity of the negative neurons.}
    \label{fig:probability_functions}
\end{figure}

To provide additional insights into the design of SFFA, we present a visualization of the two probability functions relative to FFA and our proposal. Such a visualization is provided in Figure \ref{fig:probability_functions}. It is important to remark that FFA employs a constant threshold value during its training phase. Therefore, each horizontal line illustrates the behavior of a distinct probability function. In contrast, SFFA does not require any hyper-parameter, and is independent of the total activity. Consequently, we can visualize the function across any range of activity, for which we select the range from 0 to 1 for both axes.

Upon initial inspection, our probability function exhibits a smoother transition from high values to lower ones, resulting in a broader range of in-between values. This extended domain can be leveraged to produce more accurate predictions, especially when presented with samples for which the model is uncertain. In contrast, the plot corresponding to FFA is characterized by two high-polarity regions, potentially leading to higher levels of overconfidence in the results. Another important drawback of FFA lies in its reliance on the threshold value for the function to properly behave as a probability estimator. Low threshold values are incompatible with achieving sufficiently low probability estimates, thereby hindering the model's capacity to properly distinguish between positive and negative samples. To achieve low enough probability values, the threshold needs to be set higher than $2$, ensuring a minimum probability of $0.1$. However, increasing the threshold implies that models must subsequently increase their total activity to achieve sufficient activity to surpass the threshold when presented with positive samples. This can also impact on the model's performance, especially in scenarios where the model is not capable of reaching an activity above the threshold. This is the case for bounded activation functions, where a proper analysis of the threshold is required. These shortcomings in the probability function are the main motivations for the development of SFFA, as it is able to surpass them, allowing for a less constrained model design.

\section{Sparsity of SFFA over FFA}
\label{sec:sparsity_jiji}
Sparsity has been regarded as a requirement to build energy efficient systems, with spiking neural networks gaining momentum as the most prominent example of this paradigm. Empirical evidence given by \citet{tosato2023emergent} showed that networks trained using FFA create a sparse latent representation of the different classes. This effect has been further exposed in other works such as \citep{ororbia2023predictive} or \citep{ororbia2023learning}. Further along this line, a theoretical analysis of the algorithm presented by \citet{yang2023theory} proved that an unbounded ReLU-based network trained via FFA usually converges to a sparse latent representation.

\begin{figure}[h]
    \centering
    
    \includegraphics[width=\textwidth]{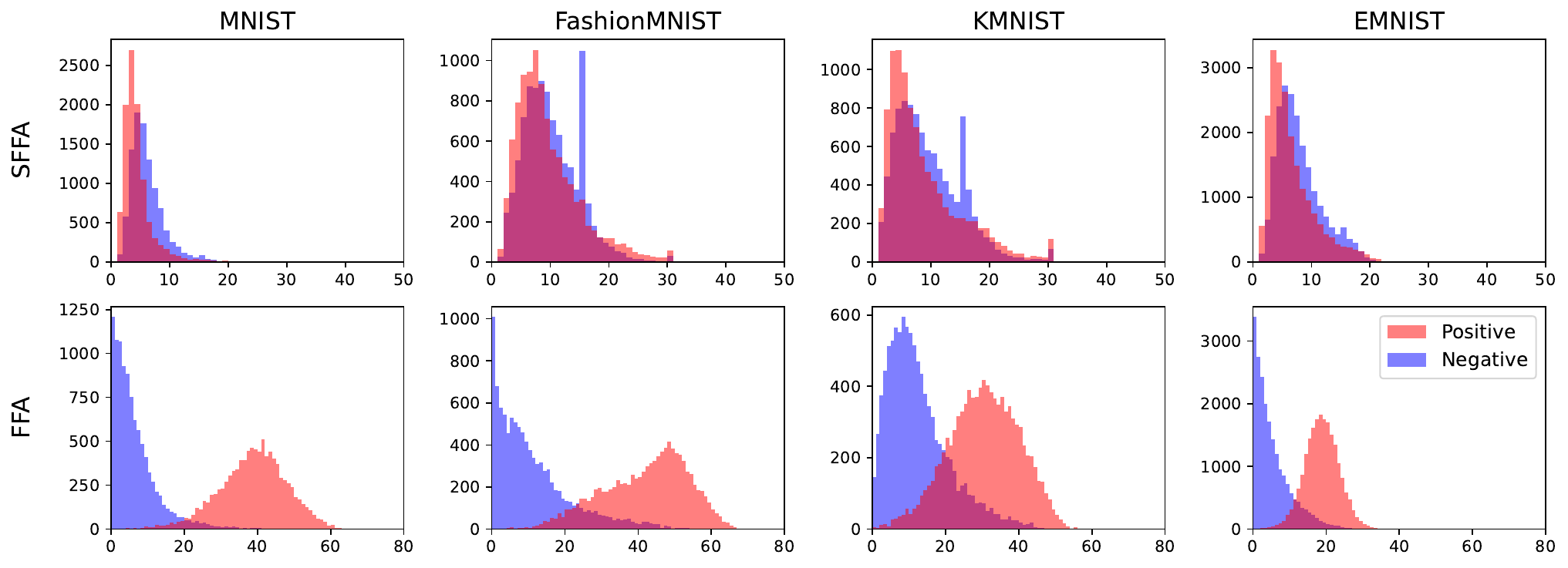}
    \caption{Distribution of neurons sharing more than $1\%$ of the latent vectors activation. The x-axis stands for the amount of active neurons, while the y-axis shows the total number of latent vectors with an specific number of high-activity neurons.}
    \label{fig:sparsity}
\end{figure}

Given this well-known sparsity effect, in this appendix we provide empirical evidence on the sparsity of the latent space of networks trained using our SFFA. To this end, Figure \ref{fig:sparsity} depicts the distribution of the number of highly active neurons in the latent space of the neural network trained via FFA/SFFA over each dataset considered in the experiments. Due to the different distribution between positive and negative latents in FFA, we highlight each distribution with a different color. Additionally, in Table \ref{tab:hoyer_tab} we present the average Hoyer score of the latent spaces, which is a well-known sparsity metric in which values closer to $1$ imply higher sparsity \citep{hurley2009comparing}.

It becomes evident upon initial inspection that our SFFA model results in sparser latent distributions, having fewer than 20 highly active neurons in most instances. Additionally, it can also be noted that positive and negative samples share similar distribution of active neurons, probably due to the symmetry in the behavior of both sets during training. While initially we advocated for the use of a k-WTA dynamics, the total active neurons usually lie far beneath this number, with only a small subsample in FashionMNIST and KMNIST reaching this range of activity. However, our side experiments also revealed that whenever this lateral inhibition dynamic was not set, a large number of neurons would specialize in the same class, resulting in a highly distributed activity over the layer. Therefore, this inhibition dynamics serve as an early regularizer of the networks, ensuring the sparsity of the model. One clear argument justifying the requirement of k-WTA dynamics can be observed in the distributions of FashionMNIST and KMNIST, where a abrupt spike in the count arises around $15$ active neurons. This spike arises from a less sparse representation in negative samples with high confidence, where negative samples only have activity in negative neurons, but employ the designed 15 k-winner neurons.

A different picture can be seen in the latent distributions of FFA networks, which clearly exhibit a less sparse latent distribution. As expected, negative samples remain sparser, as they are expected to have reduced activity, which naturally results in more inactive neurons and ultimately, in more sparse states. Generally, positive states in FFA can reach over 30 highly active neurons, with only EMNIST eliciting a coarser distribution. Furthermore, when examining the Hoyer scores of both models in Table \ref{tab:hoyer_tab}, latent spaces of networks trained via SFFA achieve scores over $0.97$ for positive samples, while those corresponding to FFA barely surpasses $0.9$ only for EMNIST.
\begin{table}[h]
    \centering
    \small
    \caption{Average Hoyer coefficient computed over the latent vectors of the test split of each dataset. Values closer to $1$ indicate higher sparsity.}
    \begin{tabular}{cccccc}
         \toprule
         Algorithm & Instance type & MNIST & FMNIST & KMNIST & EMNIST \\
         \midrule
         \multirow{2}{*}{SFFA} & Positive & 0.990 & 0.972 & 0.976 & 0.986 \\
         & Negative & 0.984 & 0.968 & 0.972 & 0.980 \\
         
         \multirow{2}{*}{FFA} & Positive & 0.855 & 0.836 & 0.884 & 0.924 \\
         & Negative & 0.978 & 0.959 & 0.951 & 0.984 \\
         
         \bottomrule
    \end{tabular}
    \label{tab:hoyer_tab}
\end{table}

\begin{figure}[h]
    \centering
    
    \includegraphics[width=\textwidth]{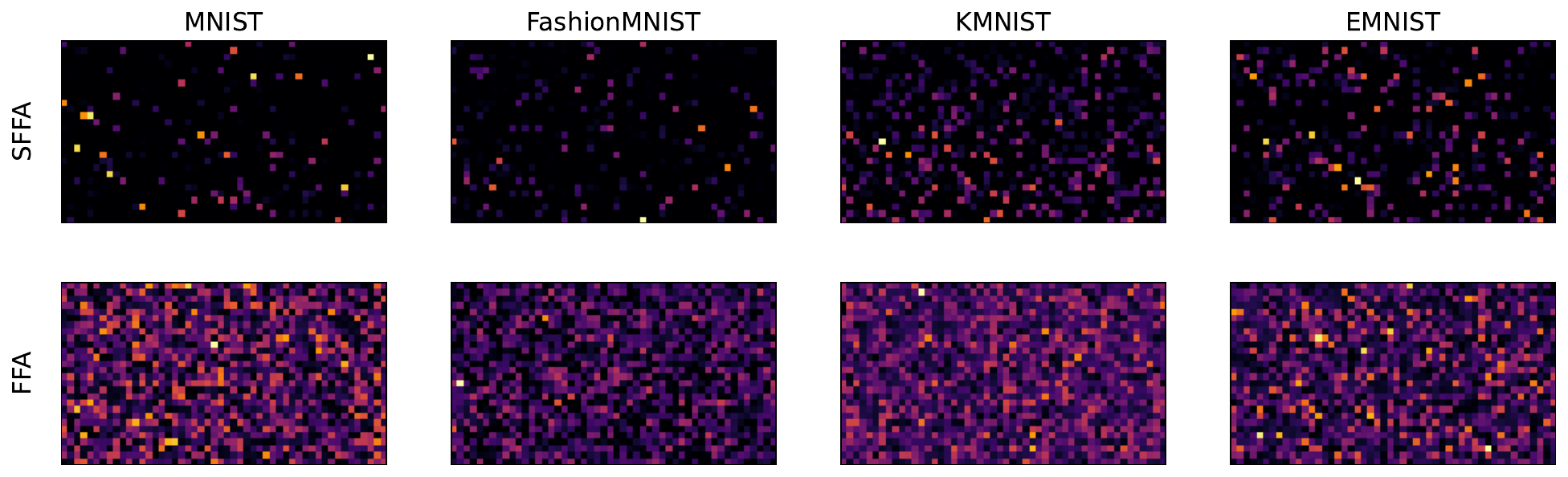}
    \caption{Aggregated latent activations of the first layer over each test dataset for SFFA (first row) and FFA (second layer). The latent activation vector has been reshaped to a 28 by 50 rectangle for the sake of readability. The color of each neuron displays its activity, with darker colors representing less activity across the dataset at hand.}
    \label{fig:sparsity2}
\end{figure}

In addition to the sparsity of the latent vectors, it is important to analyze the degree of usage of each neurons over the whole dataset when training the network with both algorithms. For instance, unused neurons over the original dataset could be employed to learn novel concepts while minimally impacting on previously gained knowledge. Although sparse latent distributions ease the existence of these \textit{dead} neurons, they do not guarantee them, which is the case of FFA. This statement is supported by Figure \ref{fig:sparsity2}, where we depict the aggregated activity of the latent vectors over the whole dataset. We can observe in this figure that networks trained using FFA tend to employ all the neurons homogeneously, creating sparse representations which encompass the whole set of output neurons. By contrast, networks trained via SFFA also exhibit sparse neural usage patterns, with a few neurons comprising almost all the activity of the latent space. The clearest example of this effect is given in the MNIST dataset: FFA distributes the latent activity over all neurons, while SFFA only employs a reduced set of highly active neurons.

\section{Gradient and Update Behavior} \label{ap:gradient}

\newcommand{\eRara}{\|\mathbf{l}_{\mathcal{E}}\|^2}
\newcommand{\iRara}{\|\mathbf{l}_{\mathcal{I}}\|^2}

\newcommand{\mAe}{\mA^{\mathcal{E}}}
\newcommand{\mAi}{\mA^{\mathcal{I}}}

This Appendix provides additional theoretical insights into the learning dynamics of the SFFA, its claimed instability and the inclusion of a regularization term made in Expression \ref{eq:loss_regu}.

One fundamental property of Forward-Forward-like algorithms is the locality of updates, i.e., training each layer is based exclusively on the information contained within it. This clashes with back-propagated networks, where a layer's update is based on the gradient information carried from all subsequent layers. This inter-dependence between layer updates significantly increases the complexity of one update. 

We specifically focus on weight and activity updates of models trained with SFFA using a single positive sample. We can omit adding a negative sample to our characterization study due to the symmetry of SFFA, by which updates on these samples exhibit the same behavior, only with the positive and negative sets reversed. Similarly, we will only examine the effects on networks with a single layer, as the training dynamics result in equal behavior across all layers. For this rationale we simplify the notation from $\mathbf{l}_{\ell}$ to $\mathbf{l}$ for the sake of a better readability of our derivations. 

While our experiments consider sigmoid activation functions, the complexity introduced by this non-linearity results in more difficult mathematical developments. To overcome this issue, differently from the experiments reported in the paper (which consider sigmoid activation functions), we assume ReLU activation functions for the rest of the Appendix as they allow for a more tractable analysis and still leverage several key aspects of the initial formula. Similarly, we purposely avoid using bias terms in our equations to reduce the complexity of the expressions. 

Recapping on Subsection \ref{sec:methodology}, SFFA relies on a modified BCE loss with a custom probability function related to the goodness score given in Expression (\ref{eq:probability_function}). Therefore, under the assumptions made in the previous paragraph, we end up with a loss expression given by:
\begin{equation}
    \mathcal{L} = -\log\left(p(\mathcal{E}, \mathcal{I})\right) = -\log\left(\frac{\eRara}{\eRara + \iRara}\right),
\end{equation}
where $\mathcal{E}$ refers to the output of the positive set of neurons and $\mathcal{I}$ refers to the output of the negative set of neurons. We adhere to the convention of $\|x\|$ representing the standard Euclidean norm. Given this loss function, a straightforward computation leads to the change direction of each weight of the layer expressed as:
\begin{equation}
\label{eq:weight_change}
    \frac{\partial \mathcal{L}}{\partial w_{ij}} = 
- \underbrace{\frac{\iRara}{\eRara}}_{\text{Ratio } R} \underbrace{\frac{1}{\eRara+\iRara}}_{\text{Inverse Sum  } S} \left(2 f\left(A_j\right) \frac{\partial f(A_j)}{\partial A_j} x_i \right)
\end{equation}
where $f : \mathbb{R} \rightarrow \mathbb{R}$ denotes an almost-everywhere differentiable function, $A_j$ refers to the pre-activation value of neuron $j$, and $x_i$ denotes the value of input $x$ in coordinate $i$. As the majority of algebraic manipulations in this Appendix are solely based on this update, two key elements in it will be simplified to yield new variables: the ratio of negative to positive neurons (denoted by $R$); and the inverse of the sum of squared activity (referred to as $S$).

Given this weight gradient, we can compute the exact formula of the change in activity of each neuron after a learning step. To this end, we will denote the change in the pre-activity of neuron $j$ in the output layer as $\Delta A_j$. Similarly, recalling from the stochastic gradient descent algorithm, we define a learning rate hyper-parameter denoted by $\alpha$. Consequently, the activity change is equal to:
\begin{equation}
    \begin{aligned}
        \Delta A_j =& \sum_i \left( w_{ij} + 2 \alpha \frac{\iRara}{\eRara} \frac{1}{\eRara+\iRara}  A_j x_i \right) x_i  - \sum_i w_{ij}x_i & \\
        &=\sum_i \left(2 \alpha \frac{\iRara}{\eRara} \frac{1}{\eRara+\iRara}  A_j x_i \right) x_i & \\ 
        &=2 \alpha A_j \frac{\iRara}{\eRara} \frac{1}{\eRara+\iRara} \sum_i x_i^2 & \\
        &=\underbrace{2 \alpha \frac{\iRara}{\eRara} \frac{1}{\eRara+\iRara} \|x\|^2}_{\text{Constant in Layer  } \gamma_1} A_j.&
    \end{aligned}
\end{equation}

It is clear from the final expression in the above formulae that the change in activity is mostly driven by a constant factor independent of index $j$, which implies that all neurons possess the same scaling factor. Similarly to Expression \ref{eq:weight_change}, we hereafter refer to this constant expression as $\gamma_1$.

By following a similar process for the negative neurons, we achieve the converse weight update formula given by:
\begin{equation} \label{eq:weight_change_neg}
\frac{\partial \mathcal{L}}{\partial w_{ij}} =
\frac{1}{\eRara+\iRara} \left(2 f\left(A_j\right) \frac{\partial f(A_j)}{\partial A_j} x_i \right).
\end{equation}

Clearly, Equation \ref{eq:weight_change} and Equation \ref{eq:weight_change_neg} are nearly identical to each other, differing only by a scaling factor of $-R$. Consequently, the value in pre-activation activity for neurons in the negative set also differs by this scaling factor. Similarly, we abbreviate the constant term of this expression as $\gamma_2$, yielding:
\begin{equation}
    \Delta A_j = \underbrace{- 2 \alpha \frac{\|x\|^2 }{\eRara+\iRara}}_{\text{Constant in Layer  } \gamma_2} A_j.
\end{equation}

By inspecting the pre-activation update in both expressions, we can notice that both comprise a scaling factor relative to the layer and the previous neural outcome. A similar behavior was observed by \citet{hinton2022forward} in the original formulation of the FFA, where training samples would lead to a direct scaling of the output vectors. However, due to the split nature of SFFA, each neural set is scaled independently, leading to a non-linear dynamic in the layer. Nevertheless, since each update is a scaling of the previous activity with a constant that is independent of individual neurons (but dependent on the entire set of neurons), we can generalize the expression for the whole layer vector $\mA$. Under this notation, we will refer to $\mA^{\mathcal{E}}$ and $\mA^{\mathcal{I}}$ as the positive and negative sets of neurons, respectively. Since data normalization regimes can be manually adjusted, for the sake of simplicity we assume a normalized input $\|x\|=1$. Therefore, the activity change in the layer is given by:
\begin{equation}
    \Delta \mA = \left\{\begin{aligned}
    \Delta \mA_{\mathcal{E}} = \gamma_1 \mA_{\mathcal{E}} = 2\alpha R S \mA_{\mathcal{E}} \\ 
    \Delta \mA_{\mathcal{I}} = \gamma_2 \mA_{\mathcal{I}} = -2\alpha S \mA_{\mathcal{I}}
\end{aligned}\right. \Longrightarrow 
\Delta \mathbf{l} = \left\{\begin{aligned}
    \Delta \mathbf{l}_{\mathcal{E}} = \gamma_1 \mathbf{l}_{\mathcal{E}} = 2\alpha R S \mathbf{l}_{\mathcal{E}} \\
    \Delta \mathbf{l}_{\mathcal{I}} = \gamma_2 \mathbf{l}_{\mathcal{I}} = -2\alpha S \mathbf{l}_{\mathcal{I}}
\end{aligned}
\right.
\end{equation}

The last implication comes as a natural conclusion due to the use of the ReLU activation function when $2\alpha S < 1$. As each pre-activity neuron is only scaled by a positive factor, the sign after the update stays the same. Hence, the ReLU function will act as a identity for the updated neurons. 

Under this formulation, we can compute the new positive and negative activity of the layer. To this end, we denote $\mathbf{l}_{\mathcal{E}} + \Delta \mathbf{l}_{\mathcal{E}}$ as the updated layer output, given the updated pre-activation activity $\mA_\mathcal{E} + \Delta \mA_{\mathcal{E}}$. From this, we obtain:
\begin{equation}
\label{eq:each_activity_change}
    \left\{ \begin{aligned}
    \|\mathbf{l}_{\mathcal{E}} + \Delta \mathbf{l}_{\mathcal{E}} \|^2 &= \sum_i \left( \mathbf{l}_{\mathcal{E},i}  + 2\alpha RS \mathbf{l}_{\mathcal{E},i} \right)^2 &=& \left(1+ 2\alpha RS\right)^2 \|\mathbf{l}_{\mathcal{E}}\|^2 \\
     \|\mathbf{l}_{\mathcal{I}} + \Delta \mathbf{l}_{\mathcal{I}}\|^2 &= \sum_i \left( \mathbf{l}_{\mathcal{I},i}  - 2\alpha S \mathbf{l}_{\mathcal{I},i} \right)^2 &=& \left(1- 2\alpha S\right)^2 \|\mathbf{l}_{\mathcal{I}}\|^2
\end{aligned}\right.
\end{equation}

By employing this change in the activity of each set, it is straightforward to verify that, given $2 \alpha S < 1$, the goodness of the layer always increases:
\begin{equation}
    \label{eq:good_increases}
    \begin{aligned}
        \frac{\eRara}{\eRara+\iRara} \leq \frac{\eRara(1+\gamma_1)^2}{\eRara (1+\gamma_1)^2 + \iRara (1+\gamma_2)^2} & \Leftrightarrow \\  
    \eRara (1+\gamma_1)^2 + \iRara (1+\gamma_2)^2 \leq (1+\gamma_1)^2 \eRara+(1+\gamma_1)^2 \iRara & \Leftrightarrow \\
    (1+\gamma_2)^2 \leq (1+\gamma_1)^2, & 
    \end{aligned}
\end{equation}
which holds true as $\gamma_2^2 \leq 1 \leq \gamma_1^2$.

\subsection{Instability of the Gradient Step}
\label{ap1:inestability}

Our derivations above conclude that the updates in the weights made by SFFA are driven by two simple formulas, in which most terms are variables relative to the activity of the neuron sets. However, these scaling factors are not naturally bounded, which can result in arbitrarily large updates. This instability in the gradients can lead to unstable learning dynamics, with models making abrupt weight changes that negatively affect the performance over the rest of the dataset. 

The weight gradient can diverge because of two terms, $R$ and $S$. The first term, $R$, refers to the ratio of the negative and positive activity. This term is constrained within the range $[0, 1]$, as positive samples are expected to achieve higher positive scores than negatives ones. However, this condition is not guaranteed, as outliers may emerge during training, which can even lead to $\eRara=0$, causing an error during learning. To mitigate this effect, we employ a clamping strategy to set a minimum and maximum achievable goodness value. 

The second unstable term, $S$, refers to the inverse of the layer activity. In order for this term to become unstable, the activity of a latent vector must come close to zero. Under this scenario, the update can be proven to diverge, as:
\begin{equation}
    \Delta \mathbf{l}_{\mathcal{E}, j} \sim \frac{\iRara}{\eRara} \frac{\mathbf{l}_{\mathcal{E}, j}}{\eRara+\iRara} \leq \frac{\mathbf{l}_{\mathcal{E}, j}}{\eRara+\iRara} \leq \frac{\mathbf{l}_{\mathcal{E}, j}}{\mathbf{l}_{\mathcal{E}, j}^2+\iRara} \leq \frac{\mathbf{l}_{\mathcal{E}, j}}{\mathbf{l}_{\mathcal{E}, j}^2}\leq \frac{1}{\mathbf{l}_{\mathcal{E}, j}} \stackrel{\mathbf{l} \rightarrow 0}{\longrightarrow} \infty
\end{equation}

To avoid this diverging behavior, we propose to include the regularization term in the loss function presented in Equation \ref{eq:loss_regu}, which effectively pushes the overall activation away from low-activity regions. Additionally, as our goodness function does not depend on the total activity of the layer, changing the activity of the output should not impact the model's performance. Similarly, within regions far from the origin, the impact of the regularization term should become negligible, thereby not affecting the gradient. 

The design of the regularization term has been developed to push the latent space from regions close to zero, but, at the same time, to have little to no impact on the loss function in regions with high activity. The exponential function $e^{-\alpha \|\mathbf{L_{\ell}}\|_n}$ satisfies these properties, while being computationally cheap. The value $n$ of the norm $\|\cdot\|_n$ of the exponent and the multiplicative nature of the regularization term were chosen by selecting the best performing alternative after training several models using different configurations. However, models trained using an Euclidean norm, or models using the exponential term in an additive way resulted in almost no accuracy loss, while also fixing the low activity issue.

\begin{figure}[h]
    \centering
    \subfloat[No regularization term]{\includegraphics[width=0.47\textwidth]{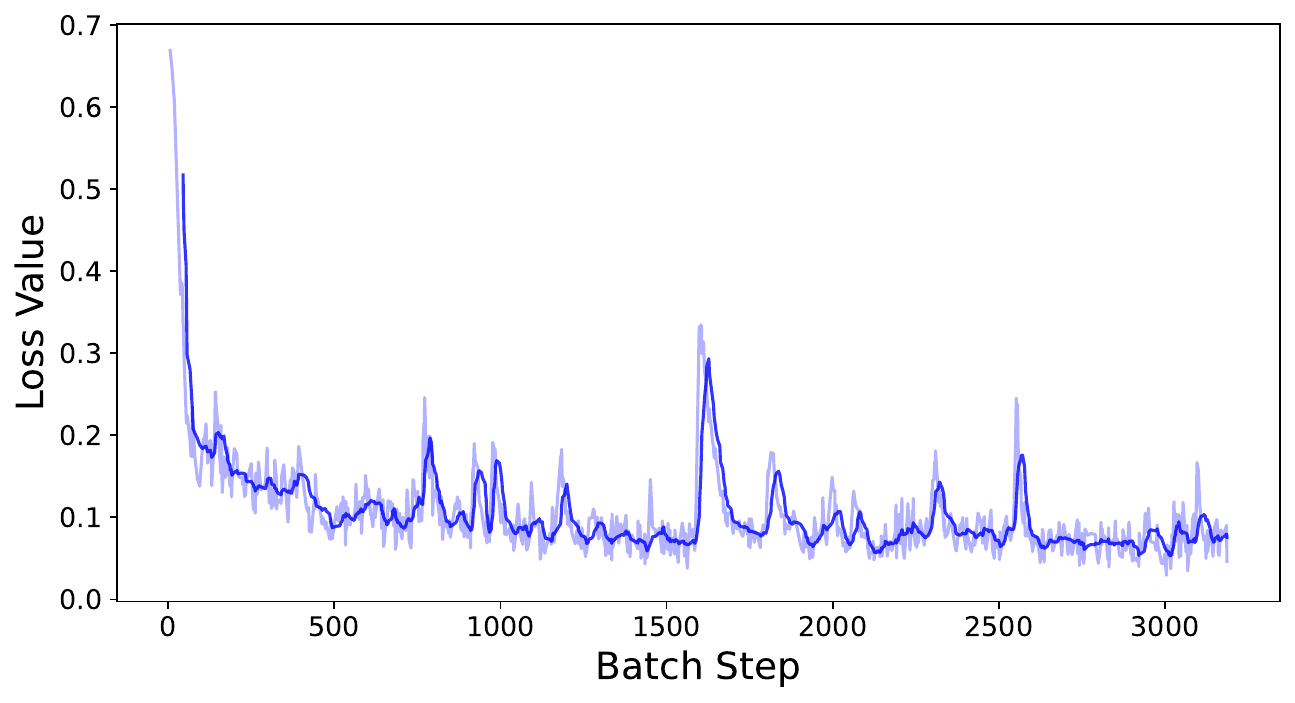}}\label{fig:linea_loss_unreg}
    \hfill
    \subfloat[Regularization term]{\includegraphics[width=0.47\textwidth]{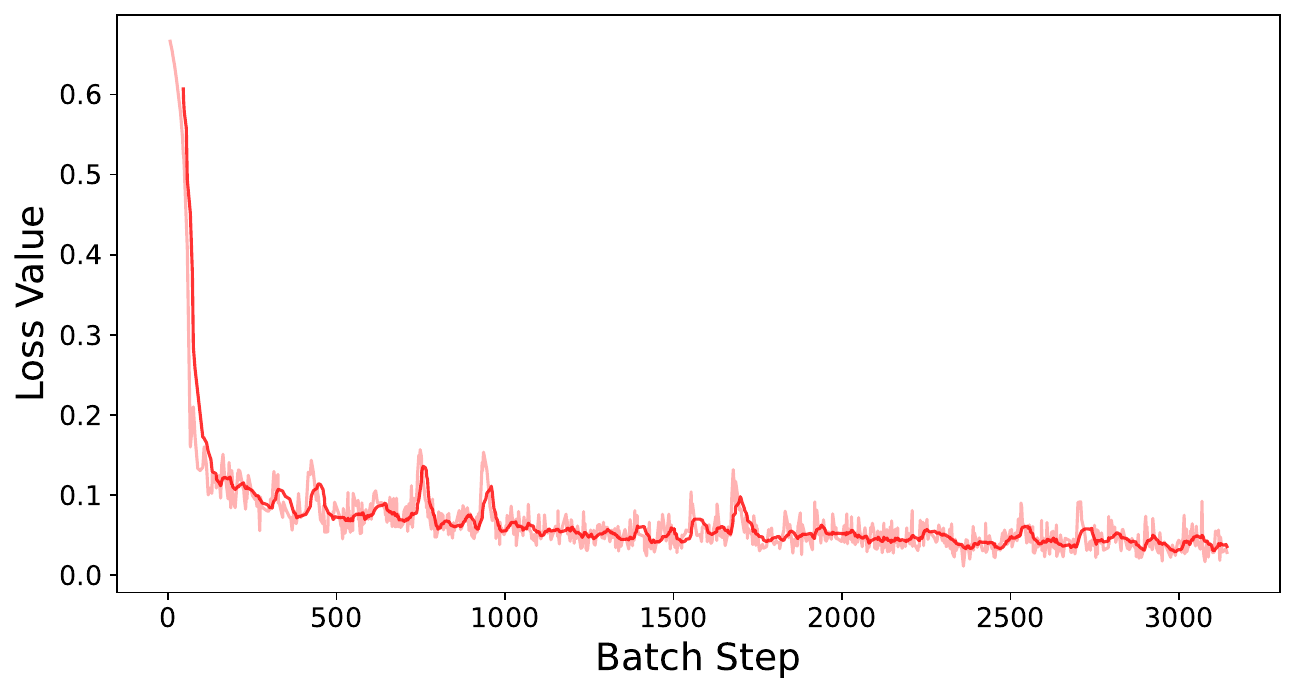}}\label{fig:linea_loss_reg}
    \caption{Convergence of the BCE loss with and without regularization of a neural network with ReLU activation over the MNIST dataset. The horizontal axis represents the number of updates (i.e. number of batches in 26 epochs).}
    \label{fig:ambas_imagenes}
\end{figure}

To empirically demonstrate the enhanced stability in the regularized loss, we train two networks using ReLU activation functions on the MNIST dataset, one model with and the other without regularization. Figure \ref{fig:ambas_imagenes} depicts the enhanced smoothness in the loss in the networks trained using the regularization terms. Sharp increases in the loss value can be observed for the non-regularized network, whereas the inclusion of the regularized term elicits a more stable behavior of the loss along the updates of the network. This buttresses the decision to implement a regularization term in the loss of the proposed SFFA.

\subsection{Total Activity after a Change}
\label{ap1:total_activity}

One significant conclusion that results from the scaling dynamics in the activity change is the constant growth in the total activity which the model attains after each training step. As mentioned previously, this effect was already seen in the original formulation of the FFA, where the total activity reflected the final goodness score. However, due to the $\mathcal{O}(1)$ behavior of our goodness function, this activity growth does not affect our goodness score. However, as has been proven in Subsection \ref{ap1:inestability}, the increase in the total activity of the layer contributes to the stability of the layer. Essentially, this can be proved by comparing the activity before and after the update, given by:
\begin{equation}
    \begin{aligned}
    \label{eq:total_activity_increase}
        \eRara + \iRara &< (1+2\alpha RS)^2\eRara + (1-2\alpha S)^2 \iRara &\Leftrightarrow \\
        0 &< 4\alpha \frac{\iRara}{\eRara} S \eRara + 4 \alpha^2R^2S^2 \eRara - 4\alpha S \iRara + 4\alpha^2S^2\iRara &\Leftrightarrow \\
        0 &< 4\alpha^2 R^2S^2\eRara + 4\alpha^2S^2\iRara &\Leftrightarrow \\
        0 &< 4\alpha^2S^2\iRara\left(1 + R\right)&
    \end{aligned}
\end{equation}

Several elements can be canceled, while the remaining ones can be grouped into a single factor. The last inequality holds, as all the elements on the right-hand side of the expression are positive. Additionally, we can also confirm that the inequality is strict, as the only elements that could equal zero are the learning rate or the negative activity. However, in order to learn, we require the learning rate to be a positive number. Similarly, having the negative activity to zero would imply that the model is already in an optimal configuration, resulting in no change in activity.

\subsection{Direction of the Update}
\label{ap1:direction}

We delve further into the behavior of the activity update by exploring the effect of the change within the $\mathcal{E}/\mathcal{I}$ plot. This plot represents the pair $(\eRara, \iRara)$ as coordinates in a plane, from which we can study the change in the ratio of these variables. Given the scaling nature of each neural component, we can easily define a discrete recurrent relationship, from which the tangent formula of the update is given by:
\begin{equation}
    \frac{\|\mathbf{l}_{\mathcal{I}} + \Delta \mathbf{l}_{\mathcal{I}}\|^2 - \iRara}{\|\mathbf{l}_{\mathcal{E}} + \Delta \mathbf{l}_{\mathcal{E}}\|^2 - \eRara} = 
    \frac{\iRara \left(1-2\alpha S\right)^2 - \iRara}{\eRara\left(1+2\alpha RS \right)^2 - \eRara}.
\end{equation}

Several elements of this expression can be simplified to obtain a reduced version of it, resulting in:
\begin{equation}
    \begin{aligned}
        \frac{\iRara \left(1-2\alpha S\right)^2 - \iRara}{\eRara\left(1+2\alpha RS \right)^2 - \eRara} &= 
        \frac{\iRara}{\eRara} \frac{\left(1-2\alpha S\right)^2 - 1}{\left(1+2\alpha RS \right)^2 - 1} &=\\ 
        -R\frac{4\alpha S - 4\alpha^2 S^2}{4\alpha RS+4\alpha^2 R^2S^2} 
        &= -\frac{4\alpha RS - 4\alpha^2R S^2}{4\alpha RS+4\alpha^2 R^2S^2} = -\frac{1 - \alpha S}{1+\alpha RS}.&
    \end{aligned}
\end{equation}

The final term of the expression consist only of 3 variables, from which we can easily understand its behavior. Notably, this expression naturally converges to $-1$ if $\alpha S \rightarrow 0$, proving that the update direction to higher goodness values is stable under two circumstances: low $\alpha$ values, which can be manually adjusted as an external hyper-parameter, or high total activity, which can be enforced with regularization terms in the loss. Under these situations, the direction of the update follows a linear trend towards $\mathcal{I} = 0$. When assessed together with Equation \ref{eq:total_activity_increase}, these results show that models achieve more linear updates the more they are trained on the same sample. In contrast, the value of the ratio $R$ only regulates the impact of the denominator, producing a steeper linear decay the further it is from optimal accuracy values.

Similarly to previous results, the only unstable situation in the change direction arises when $\alpha S$ reaches higher values, which can even reverse the direction of the update. However, this effect does not impact the performance of training on single samples. It occurs because when presented with a low-activity vector, it is actively pushed from regions near zero within a low number of steps. Once the vector attains a high enough activity, the value of $S$ is low enough for the direction to converge to a linear behavior, resulting in optimal activity updates. High values of $\alpha$ are not studied, as this hyper-parameter is user-defined and typically set to low values to ensure stability in the training phase.

\begin{figure}[h]
    \centering
    \includegraphics[width=0.8\textwidth]{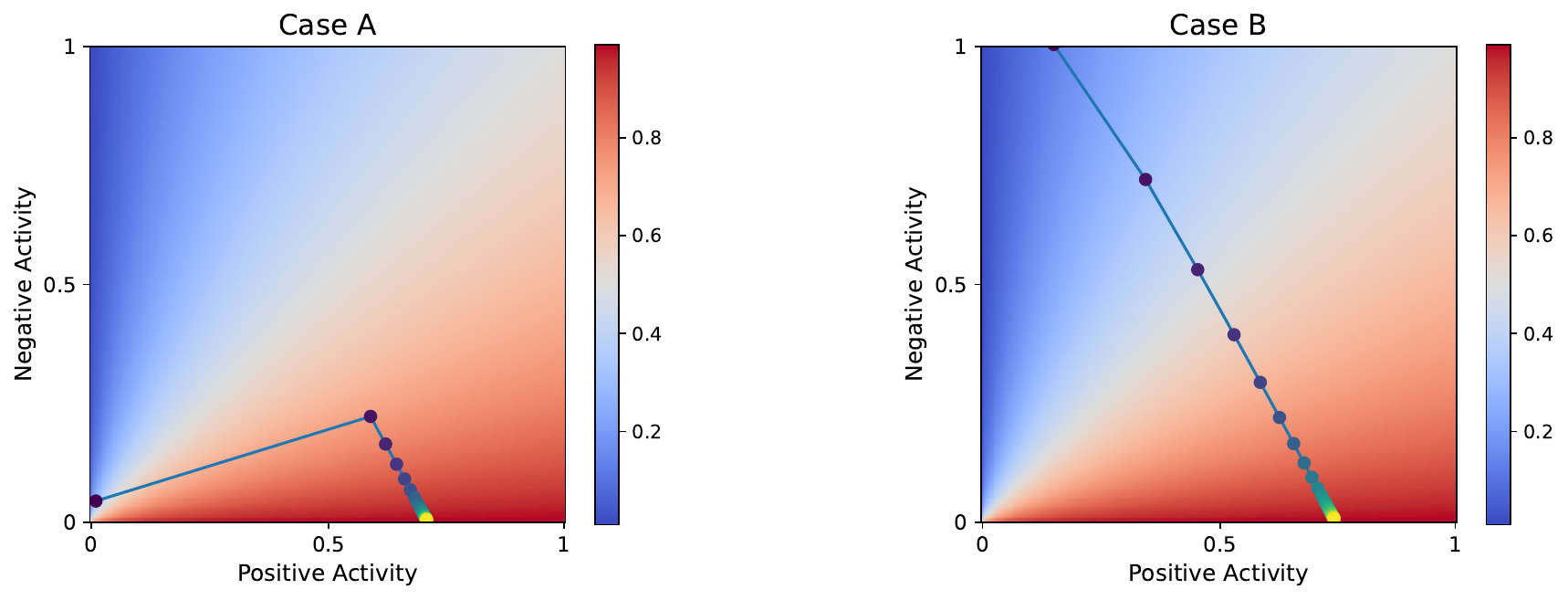}
    \caption{Simulation of the transformation in the activity values of positive and negative samples. Both simulations are executed for 20 iterations with $\alpha=0.1$. The starting point of Case A is $(0.02, 0.04)$ and that of Case B is $(0.3, 1)$.}   
    \label{fig:experimental}
\end{figure}

To further verify these theoretical results, we evaluate the behavior of two activity vectors by relying on the step equations of $\mathcal{E}$ and $\mathcal{I}$ presented in Equation \ref{eq:each_activity_change}. The results of the experiment, depicted in Figure \ref{fig:experimental}, clearly indicate that the stable behavior of these updates follows an approximate linear trend, converging to values close to $\mathcal{I} = 0$. Additionally, as shown in Case A, samples starting with low activity thresholds suffer from abrupt changes in the overall activity, getting quickly pushed into higher activity regions. After this push, the direction of the updates regularizes itself into the desired linear behavior. Additionally, we can observe the diminishing goodness increase as the simulation advances, as the total change exponentially reduces as the system gets closer to optimal values.

The optimal stability achieved by having higher activity states serves as additional motivation for the integration of the activity regularization term in the loss function in Equation \ref{eq:loss_regu}. Even though we can guarantee that within single sample scenarios the model would regularize itself, these abrupt changes can impact the performance of the model with the rest of the training data.

\section{Experimental Setup: Additional Information}
\label{ap:hyperparameters}
This Appendix extends the information about the setup considered in our experiments. Source code of SFFA and the scripts producing the results for RQ1 and RQ2 can be found in the following GitHub repository: \url{https://github.com/erikberter/SFFA_CoLLAS}.

\subsection{Performance metrics}
\label{ap:metrics_1}
To measure the quality of the network over the different experiments, we utilize three well-known CL metrics, namely, Accuracy ($Acc$), Forward-Transfer ($FwT$) and Forgetting Measure ($Fgt$). Given a stream of $T$ different training tasks, we define the model's accuracy over task $i$ after learning on experience $j$ as $\textup{acc}_{i,j}$. Additionally, we denote the average accuracy of the untrained model over  task $i$ as $\overline{\textup{acc}_i}$. Based on these definitions, the CL metrics reported in our experiments are given by:
\begin{align}
\label{eq:metrics}
Acc = \frac{1}{T} \sum^{T}_{i=1} \textup{acc}_{i, T}, \qquad FwT = \frac{1}{T-1} \sum^{T}_{i=2} \textup{acc}_{i-1, i} - \overline{\textup{acc}_{i}}, \qquad Fgt = \frac{1}{T-1} \sum^{T-1}_{i=1} \textup{acc}_{i,i} - \textup{acc}_{T,i}
\end{align}
where $T$ denotes the total number of tasks. 

\subsection{Hyper-parameter values}
\label{ap:hyperparameters_1}

In this section we detail the hyper-parameter values used for the experiments reported in the paper. Table \ref{tab:params_training} corresponds to the experiments related to RQ1, whereas Table \ref{tab:params_cl} indicates the values for the hyper-parameters that relate to the experiments of RQ2.
\begin{table}[h]
    \centering
    \caption{Hyper-parameter values for the experiments corresponding to RQ1.}
    \resizebox{\textwidth}{!}{\begin{tabular}{lp{11cm}c}
        \toprule
        Variable &  Definition & Value \\
        \midrule
        Learning rate  & Rate of change of weight updates & $10^{-4}$  \\
        Hidden layer & Number of hidden layers & $2$ \\
        Residual connections & Whether the models used residual connections from the inputs to hidden layers  & True \\
        Activity normalization & Whether the models normalize the activity in the hidden layers  & True \\
        Goodness clamp & Value clamp to avoid $\log 0$ when computing the loss  & $10^{-4}$ \\
        Batch size & Number of samples within each batch
        & $512$ \\
        FFA activation & Activation function of networks trained via FFA
        & ReLU \\
        SFFA activation & Activation function of networks trained via SFFA
        & Sigmoid \\
        Epoch & Number of training epochs & $100$ \\
        Layer size & Number of neurons at each hidden layer & $1400$ \\
        Label encoding & Technique used to encode labels into the input & Symba \\
        Pattern size & Size of the encoding pattern of the Symba strategy  & $100$ \\
        Pattern density & Probability of a position being active in the generation of the encoding vector & $0.1$ \\       
        \bottomrule
    \end{tabular}}
    \label{tab:params_training}
\end{table}

\makeatletter
\setlength{\@fptop}{0pt}
\makeatother
\begin{table}[h]
    \centering
    \caption{Hyper-parameter values for the experiments corresponding to RQ2. Unless otherwise stated, hyperparameter values of Table \ref{tab:params_training} are considered for model configuration.}
    \begin{tabular}{cclc}
        \toprule
        Variable & Algorithm & Definition & Value \\
        \midrule
        lr & - & Learning rate of the models & $10^{-3}$ \\
        Epoch & - & Number of epochs per task & $2$ \\
        
        $\lambda_{\textup{EWC}}$ & EWC & Strength parameter for the loss of previous tasks & $10^5$  \\
        $\lambda_{\textup{SI}}$ &  SI & Strength parameter for the loss of previous tasks & $10^3$ \\
        $\eps_{\textup{SI}}$ & SI & Damping parameter to avoid division by zero  & $0.1$ \\
        $N$ & Replay & Number of samples of the episodic memory & $500$ \\
        $m$ & GEM & Number of samples stored per task & $20$ \\
        $\gamma_{\textup{GEM}}$ & GEM  & Strength factor for scaling the gradient & $0.5$ \\
        $\mu_{\textup{MAS}}$ &  MAS & Penalty term in the loss & $1$ \\
        $\alpha_{\textup{MAS}}$ &  MAS & Strength parameter for the loss of previous tasks & $0.5$ \\
        \bottomrule
    \end{tabular}
    \label{tab:params_cl}
\end{table}

\section{Additional Results in RQ2}
\label{ap:additional_results_rq2}

This appendix presents additional results for the CL experiments addressed in RQ2. Table \ref{tab:mnist_fgt_fwt_results} contains the results of Forgetting and Forward Transfers in terms of the metrics presented in Equation \eqref{eq:metrics} (Appendix \ref{ap:metrics_1}).

\begin{table}[h]
    \centering
    \caption{Averaged $Fgt$ and $FwT$ metrics obtained over 10 experiments with different seeds for the Split MNIST and Split Fashion-MNIST datasets. The direction of the arrow in the name of the metric indicate that lower $Fgt$ values and higher $FwT$ values are better. The highest Forgetting and Forward Transfer result among all the methods has been marked in bold at for each dataset and CL scenario.}
    \resizebox{\textwidth}{!}{\begin{tabular}{llcccccccccccccccccccc}
        \toprule
     \multicolumn{22}{c}{\multirow{2}{*}{Split MNIST}}\\
            \\ \midrule
        && \multicolumn{6}{c}{Class IL} && \multicolumn{6}{c}{Domain IL} && \multicolumn{6}{c}{Task IL}\\
        \cmidrule{3-8} \cmidrule{10-15} \cmidrule{17-22}
        Method & & \multicolumn{2}{c}{Back-propagation} & \multicolumn{2}{c}{FFA} & \multicolumn{2}{c}{SFFA} && \multicolumn{2}{c}{Back-propagation} & \multicolumn{2}{c}{FFA} & \multicolumn{2}{c}{SFFA} && \multicolumn{2}{c}{Back-propagation} & \multicolumn{2}{c}{FFA}  & \multicolumn{2}{c}{SFFA}\\ \midrule
        Metric & & $Fgt$ $\downarrow$ & $FwT$ $\uparrow$& $Fgt$ $\downarrow$ & $FwT$ $\uparrow$ & $Fgt$ $\downarrow$ & $FwT$ $\uparrow$ && $Fgt$ $\downarrow$ & $FwT$ $\uparrow$ & $Fgt$ $\downarrow$ & $FwT$ $\uparrow$ & $Fgt$ $\downarrow$ & $FwT$ $\uparrow$ && $Fgt$ $\downarrow$ & $FwT$ $\uparrow$ & $Fgt$ $\downarrow$ & $FwT$ $\uparrow$ & $Fgt$ $\downarrow$ & $FwT$ $\uparrow$ \\
        \midrule
        Adam    && 99.17 & -15.91 & 93.47 & -16.08 & 99.26 & -15.87 && 37.73 & 3.07 & 34.37 & 2.55 & 35.74 & 2.57 && 10.94 & 0.12 & 10.97& \phantom{1}0.16 & 13.74 &  \phantom{1}0.16\\
        
        EWC     && 99.25 & -15.95 & 84.35 & -16.02 & 98.48 & \textbf{-15.77} && 32.31 & 2.97 & 24.15 & 2.40 & 29.30 & 2.87 &&  \phantom{1}1.38 & 0.12 & \phantom{1}2.47 & \phantom{1}\textbf{3.49} &  \phantom{1}6.66 &  \phantom{1}3.14\\
        SI      && 99.27 & -15.95 & 81.14 & -16.06 & 98.26 & -15.86 && 31.84 & 3.53 & 24.97 & 2.17 & 33.02 & 2.97 &&  \phantom{1}0.41 & 0.16 & \phantom{1}2.43 & -0.40 &  \phantom{1}4.30 & -0.59\\
        MAS     && 97.77 & -15.90 & 92.31 & -16.11 & 98.98 & -15.91 && 28.37 & 3.20 & 34.18 & 2.45 & 36.11 & 2.46 &&  \phantom{1}\textbf{0.03} & 0.23 & 10.80& \phantom{1}0.29 & 11.47 & -0.02\\
        Replay  && 14.23 & -15.82 & 11.33 & -15.95 & \textbf{10.91} & -15.86 &&  \phantom{1}5.86 & 3.59 &  \phantom{1}4.99 & 3.12 &  \phantom{1}\textbf{4.91} & \textbf{3.80} &&  \phantom{1}0.32 & 0.13 & \phantom{1}0.09 & \phantom{1}0.85 &  \phantom{1}0.30 &  \phantom{1}0.05\\
        GEM     && 45.63 & -15.86 & 21.52 & -15.98 & 36.68 & -15.91 && 15.85 & 2.28 & 11.76 & 1.72 & 14.02 & 2.65 &&  \phantom{1}1.10 & 0.09 & \phantom{1}1.87 & -1.41&  \phantom{1}1.44 &  \phantom{1}0.61\\
     \midrule
     \multicolumn{22}{c}{\multirow{2}{*}{Split Fashion-MNIST}}\\
            \\
     \midrule 
     && \multicolumn{6}{c}{Class IL} && \multicolumn{6}{c}{Domain IL} && \multicolumn{6}{c}{Task IL}\\
        \cmidrule{3-8} \cmidrule{10-15} \cmidrule{17-22}
        Method & & \multicolumn{2}{c}{Back-propagation} & \multicolumn{2}{c}{FFA} & \multicolumn{2}{c}{SFFA} && \multicolumn{2}{c}{Back-propagation} & \multicolumn{2}{c}{FFA} & \multicolumn{2}{c}{SFFA} && \multicolumn{2}{c}{Back-propagation} & \multicolumn{2}{c}{FFA}  & \multicolumn{2}{c}{SFFA}\\ \midrule
        Metric & & $Fgt$ $\downarrow$ & $FwT$ $\uparrow$& $Fgt$ $\downarrow$ & $FwT$ $\uparrow$ & $Fgt$ $\downarrow$ & $FwT$ $\uparrow$ && $Fgt$ $\downarrow$ & $FwT$ $\uparrow$ & $Fgt$ $\downarrow$ & $FwT$ $\uparrow$ & $Fgt$ $\downarrow$ & $FwT$ $\uparrow$ && $Fgt$ $\downarrow$ & $FwT$ $\uparrow$ & $Fgt$ $\downarrow$ & $FwT$ $\uparrow$ & $Fgt$ $\downarrow$ & $FwT$ $\uparrow$ \\
        \midrule
        Adam    && 96.98 & -15.73 & 85.09 & -15.89 & 94.64 & -14.24 && 43.35 & \phantom{1}0.86 & 40.48 &  0.26 & 44.28 & -1.50 && 10.52 & \phantom{1}0.28 & 23.34 & -4.21 & 20.53 & -9.48 \\
        
        EWC     && 96.35 & -15.66 & 87.60 & -16.22 & 93.13 & \textbf{-13.69} && 43.06 & -0.71 & 33.95 &  \phantom{1}0.73 & 37.92 & -1.23 && \phantom{1}4.31 & \phantom{1}0.23 &  \phantom{1}4.63 &  \phantom{1}0.28 &  \phantom{1}5.23 & \phantom{1}\textbf{0.34} \\
        SI      && 95.99 & -15.96 & 83.64 & -15.86 & 92.23 & -14.71 && 42.44 &  \phantom{1}1.53 & 34.37 & -0.93 & 42.69 & -1.46 && \phantom{1}0.76 & -0.07 &  \phantom{1}7.57 & -8.87 &  \phantom{1}8.69 & -6.16 \\
        MAS     && 94.13 & -16.00 & 83.97 & -16.12 & 94.49 & -14.69 && 39.25 &  \phantom{1}\textbf{2.51} & 40.21 &  \phantom{1}0.11 & 42.44 & -0.25 && \phantom{1}0.49 & \phantom{1}0.25 & 21.12 & -3.94 & 18.40 & -1.85 \\
        Replay  && 18.66 & -15.63 & 18.53 & -16.88 & \textbf{18.35} & -14.86 &&  \phantom{1}\textbf{9.43} & -2.87 & 10.03 & -2.44 &  \phantom{1}9.62 & -3.64 && \phantom{1}0.66 & -0.10 & \textbf{-0.06} & -6.02 &  \phantom{1}0.32 & -6.02 \\
        GEM     && 45.78 & -15.95 & 30.08 & -15.84 & 39.86 & -14.68 && 19.91 & -1.79 & 16.37 & -2.49 & 20.18 & -2.04 && \phantom{1}1.42 & \phantom{1}0.08 &  \phantom{1}1.55 & -5.60 &  \phantom{1}1.83 & -3.10 \\
     \bottomrule
    \end{tabular}}    
    \label{tab:mnist_fgt_fwt_results}
\end{table}

\end{document}

%% file: math_commands.tex
\usepackage{amsmath,amsfonts,bm}

\def\eqref#1{equation~\ref{#1}}

\def\1{\bm{1}}

\def\eps{{\epsilon}}

\def\mA{{\bm{A}}}

\DeclareMathAlphabet{\mathsfit}{\encodingdefault}{\sfdefault}{m}{sl}
\SetMathAlphabet{\mathsfit}{bold}{\encodingdefault}{\sfdefault}{bx}{n}

